\documentclass[acmtog]{acmart}
\acmSubmissionID{442}

\usepackage{hyperref}
\usepackage{epsfig}
\usepackage{graphicx}
\usepackage{amsmath}
\usepackage{ifthen}
\usepackage{float}
\usepackage{morefloats}
\usepackage{alltt}
\usepackage{amsmath}
\usepackage{amsthm}
\usepackage{ulem}
\usepackage{mathrsfs}
\usepackage[utf8]{inputenc}
\usepackage{newlfont} 
\usepackage{floatflt}
\usepackage{wrapfig}

\usepackage[ruled]{algorithm2e} 

\SetAlFnt{\small}
\SetAlCapFnt{\small}
\SetAlCapNameFnt{\small}
\SetAlCapHSkip{0pt}

\usepackage{booktabs} 
\usepackage{gensymb}
\usepackage{subfig} 
\usepackage{multirow}
\usepackage{cleveref}
\Crefname{equation}{Eq.}{Eqs.}
\Crefname{figure}{Fig.}{Figs.}
\Crefname{section}{Sec.}{Sec.}
\usepackage[T1]{fontenc} 
\usepackage{etoolbox}
\usepackage{xfrac}
\usepackage{booktabs} 
\usepackage[export]{adjustbox}
\usepackage{algpseudocode}
\usepackage[acronym,nomain]{glossaries}
\usepackage{adjustbox}
\usepackage{xcolor}
\usepackage{wrapfig}
\usepackage[percent]{overpic}
\usepackage{soul}
\usepackage{paralist}
\usepackage{pict2e}
\usepackage{xcolor,colortbl}

\newcommand{\wide}{\textbf{W}\xspace}

\newcommand{\tele}{\textbf{T}\xspace}

\newcommand{\hlc}[2][yellow]{{%
    \colorlet{foo}{#1}%
    \sethlcolor{foo}\hl{\textbf{#2}}}%
}

\definecolor{cellyellow}{rgb}{1,1, 0.6}
\definecolor{cellorange}{rgb}{1, 0.8, 0.6}
\definecolor{cellred}{rgb}{1, 0.6, 0.6}

\newif\ifdrafting
\draftingtrue 
\ifdrafting    
    \newcommand{\ds}[1]{{\leavevmode\color[rgb]{1,0.5,0}[Deqing: #1]}}
    \newcommand{\cih}[1]{{\leavevmode\color[rgb]{0,0.4,0}[Charles: #1]}}
    \newcommand{\ckliang}[1]{{\leavevmode\color[rgb]{0,0.4,0.4}\hlc{[CK: #1]}}}
    \newcommand{\yichang}[1]{{\leavevmode\color[rgb]{0,0,1}[YC: #1]}}
    
    \newcommand{\wslai}[1]{{\leavevmode\color[rgb]{0,0,1}[WS: #1]}}
    \newcommand{\abbywu}[1]{{\leavevmode\color[rgb]{0,0.5,0}[AW: #1]}}
    \newcommand{\mk}[1]{{\leavevmode\color[rgb]{0,0.5,1}[MK: #1]}}
    \newcommand{\TODO}[1]{\textcolor{red}{[TODO] #1}}
    \newcommand{\todo}[1]{\textcolor{red}{[TODO] #1}}
\else
    \newcommand{\ds}[1]{}
    \newcommand{\cih}[1]{}
    \newcommand{\ckliang}[1]{}
    \newcommand{\yichang}[1]{}    
    \newcommand{\wslai}[1]{}
    \newcommand{\abbywu}[1]{}
    \newcommand{\mk}[1]{}
    \newcommand{\TODO}[1]{}
    \newcommand{\todo}[1]{}
\fi

\newcommand{\ignore}[1]{}
\renewcommand{\subparagraph}[1]{\vspace{1mm} \noindent\textit{\underline{{#1}}}}
\newcommand{\dataset}[1]{\textsc{#1}}

\newcommand{\ie}{\emph{i.e.}}
\newcommand{\eg}{\emph{e.g.}}

\def\refimage{I_{\text{ref}}}

\def\srcimage{I_{\text{src}}}
\def\srcimageluma{Y_{\text{src}}}

\def\srcimagelumaresampledpatch{P_{\text{src}}}
\def\srcimagelumaresampledpatchmean{\mu_{\text{src}}}
\def\srcimagelumaresampledpatchvariance{\sigma^2_{\text{src}}}

\def\warpedrefimage{{\tilde{I}}_{\text{ref}}}
\def\warpedrefimageluma{{\tilde{Y}}_{\text{ref}}}
\def\warpedrefimagelumaresampled{{\tilde{Y}}_{\text{ref}\downarrow}}
\def\warpedrefimagelumaresampledpatch{{\tilde{P}}_{\text{ref}}}
\def\warpedrefimagelumaresampledpatchmean{\mu_{\text{ref}}}

\def\targetimage{I_{\text{target}}}

\def\forwardflow{F_{\text{fwd}}}
\def\backwardflow{F_{\text{bwd}}}

\def\occlusionmask{\mathbf{M}_{\text{occ}}}
\def\trainingocclusionmask{\hat{\mathbf{M}}_{\text{occ}}}
\def\availabilitymask{\mathbf{M}_{\text{valid}}}
\def\defocusmask{\mathbf{M}_{\text{defocus}}}
\def\blendingmask{\mathbf{M}_{\text{blend}}}

\def\rejectionmask{\mathbf{M}_{\text{reject}}}
\def\flowconfmask{\mathbf{M}_{\text{flow}}}

\def\vggloss{\mathcal{L}_{\text{vgg}}}
\def\colorloss{\mathcal{L}_{\text{brightness}}}
\def\contextualloss{\mathcal{L}_{\text{cx}}}
\def\finalloss{\mathcal{L}_{\text{final}}}
\def\vgglossweight{w_{\text{vgg}}}
\def\contextuallossweight{w_{\text{cx}}}
\def\colorlossweight{w_{\text{brightness}}}

\def\modeloutputimage{I_{\text{fusion}}}
\def\modeloutputimageluma{Y_{\text{fusion}}}
\def\finaloutputiamge{I_{\text{final}}}

\def\warp{\mathbb{W}}

\newcommand{\uncrop}[1]{\textup{uncrop}(#1)}
\newcommand{\sigmoid}[1]{\textup{sigmoid}(#1)}
\newcommand{\newmax}[1]{\textup{max}(#1)}
\newcommand{\onenorm}[1]{||#1||_1}
\newcommand{\twonorm}[1]{||#1||^2_2}

\def\imagewidth{W}

\newcommand{\figref}[1]{Fig.~\ref{fig:#1}}
\newcommand{\tabref}[1]{Table~\ref{tab:#1}} 
\newcommand{\eqnref}[1]{Eq.~\ref{eq:#1}}
\newcommand{\secref}[1]{Sec.~\ref{sec:#1}}

\setcopyright{rightsretained}
\begin{document}

\acmJournal{TOG}
\acmYear{2023}
\acmVolume{42}
\acmNumber{6}
\acmArticle{}
\acmMonth{12}
\acmPrice{15.00}
\acmDOI{10.1145/3618362}

\normalem


\title{Efficient Hybrid Zoom using Camera Fusion on Mobile Phones}

\author{Xiaotong Wu}
\email{abbywu@google.com}
\author{Wei-Sheng Lai}
\email{wslai@google.com}
\author{YiChang Shih}
\email{yichang@google.com}
\author{Charles Herrmann}
\email{irwinherrmann@google.com}
\author{Michael Krainin}
\email{mkrainin@google.com}
\author{Deqing Sun}
\email{deqingsun@google.com}
\author{Chia-Kai Liang}
\email{ckliang@google.com}
\affiliation{%
 \institution{Google}
 \country{USA}
}
\renewcommand\shortauthors{Wu, et al}

\begin{abstract}
DSLR cameras can achieve multiple zoom levels via shifting lens distances or swapping lens types. 
However, these techniques are not possible on smartphone devices due to space constraints.
Most smartphone manufacturers adopt a hybrid zoom system: commonly a Wide (\wide) camera at a low zoom level and a Telephoto (\tele) camera at a high zoom level. 
To simulate zoom levels between \wide and \tele, these systems crop and digitally upsample images from \wide, leading to significant detail loss.
In this paper, we propose an efficient system for hybrid zoom super-resolution on mobile devices, which captures a synchronous pair of \wide and \tele shots and leverages machine learning models to align and transfer details from \tele to \wide. 
We further develop an adaptive blending method that accounts for depth-of-field mismatches, scene occlusion, flow uncertainty, and alignment errors.
To minimize the domain gap, we design a dual-phone camera rig to capture real-world inputs and ground-truths for supervised training.
Our method generates a 12-megapixel image in 500ms on a mobile platform and compares favorably against state-of-the-art methods under extensive evaluation on real-world scenarios. 
\end{abstract}

%
%

\ccsdesc[500]{Computing methodologies}
\ccsdesc[400]{Artificial intelligence}
\ccsdesc[300]{Computer vision}
\ccsdesc[200]{Image and video acquisition}
\ccsdesc[100]{Computational photography}

%
%

\keywords{hybrid zoom, dual camera fusion, deep neural networks} 
\maketitle

\section{Introduction}

Being able to adjust the field-of-view (FOV) (\ie, zooming in and out) is one of the most basic functionalities in photography, yet despite their ubiquity, smartphones still struggle with zoom.
Zoom lenses used by DSLR cameras require a large assembly space that is typically impractical for smartphones. 
Recent smartphones employ hybrid optical zoom mechanisms consisting of cameras with different focal lengths, denoted as \wide and \tele, with short and long focal lengths, respectively. 
When a user zooms, the system upsamples and crops \wide until the FOV is covered by \tele. 
However, almost all forms of upsampling (bilinear, bicubic, etc.) lead to varying degrees of objectionable quality loss.
The growing demand for higher levels of zoom in smartphones has led to higher focal length ratios between \tele and \wide, typically 3-5$\times$, making detail loss an increasingly important problem.

Single-image super-resolution (SISR) adds details to images but is inappropriate for photography due to its tendency to hallucinate fake content. 
Instead, reference-based super-resolution (RefSR) aims to transfer real details from a reference image. 
A variety of sources for the reference images have been explored, \eg, images captured at a different time or camera position, or similar scenes from the web.
The hardware setup in recent smartphones provides a stronger signal in the form of \wide and \tele captures. 
Recent works have thus focused on using the higher zoom \tele as a reference to add real details back to the lower zoom \wide.

Commercial solutions exist~\cite{huaweip40,HM4-DxO} but neither technical details nor datasets are publicly available. 
Academic solutions~\cite{trinidad2019multi,wang2021dual,zhang2022self} provide insights into the problem but are not practical for real-world applications. 
Specifically, these methods tend to be inefficient on mobile phones, are vulnerable to imperfections in reference images, and may introduce domain shifts between training and inference. 
In this work, we investigate these three issues and propose a hybrid zoom super-resolution (HZSR) system to address them.

\begin{figure}[t!]
    \centering
    \includegraphics[width=1.0\columnwidth]{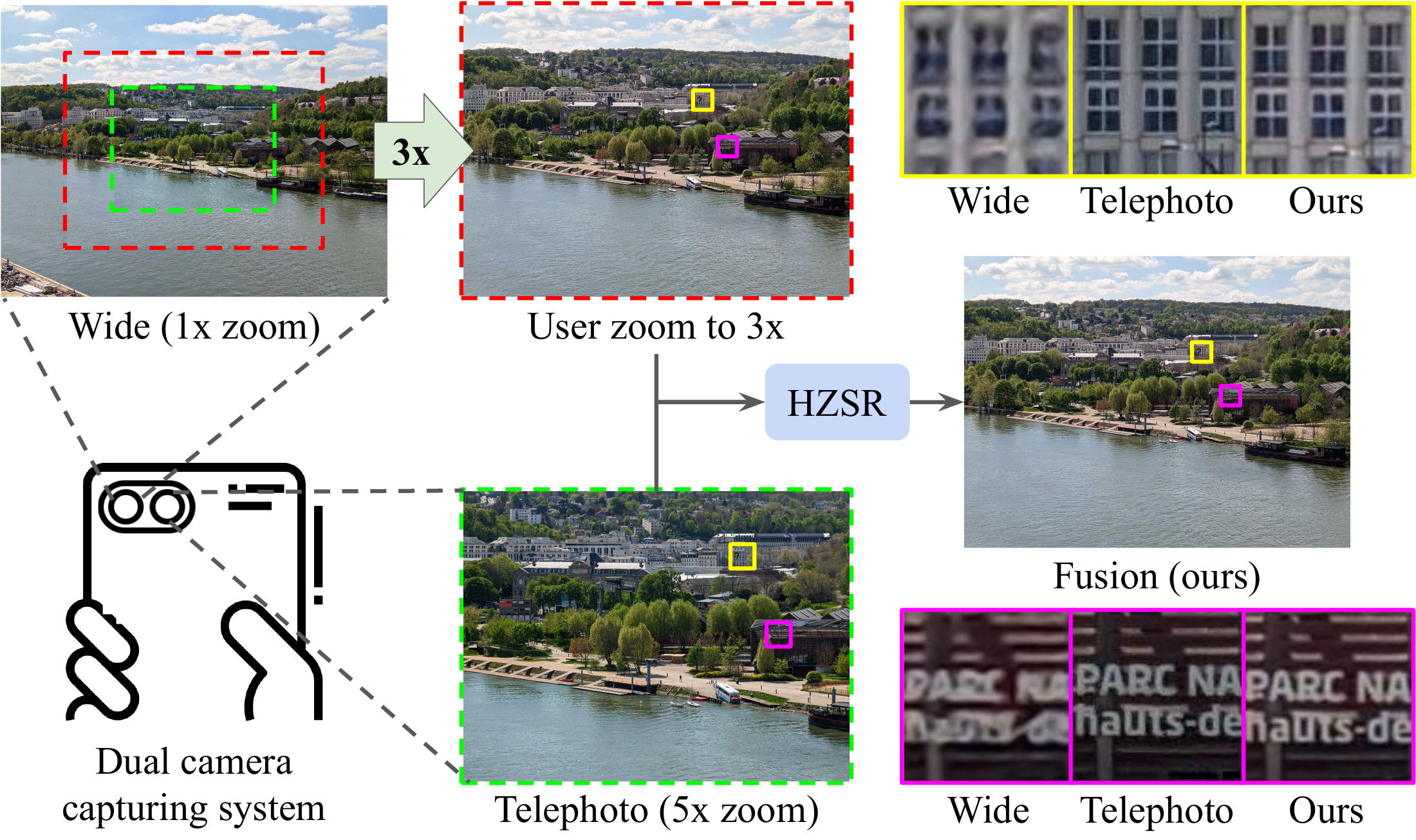}
    \caption{
        \textbf{Detail improvements in hybrid zoom.}
        The red dotted lines mark the FOV of $3\times$ zoom on $1\times$ wide (\wide) camera, while the green dotted lines mark the FOV of $5\times$ telephoto (\tele) camera.
        Image quality at an intermediate zoom range suffers from blurry details from single-image super-resolution~\cite{romano2016raisr}. 
        Our mobile hybrid zoom super-resolution (HZSR) system captures a synchronous pair of \wide and \tele and fuses details through efficient ML models and adaptive blending.
        Our fusion results significantly improve texture clarity when compared to the upsampled \wide. 
    }
    \label{fig:teaser}
\end{figure}

\paragraph{Efficient processing on mobile devices}
Existing methods require large memory footprints (\eg, out-of-memory for 12MP inputs on a desktop with an A100 GPU) and long processing times unsuitable for mobile phones. 
We develop efficient Machine Learning (ML) models to align \tele to \wide using optical flow and fuse the details at the pixel level using an encoder-decoder network. 
Our models are optimized to process 12MP inputs efficiently on mobile system-on-a-chip (SoC) frameworks, taking only 500ms extra latency and 300MB memory footprint.

\begin{figure}
    \centering
    \footnotesize
    \renewcommand{\tabcolsep}{1pt} 
	\renewcommand{\arraystretch}{0.9} 
	\renewcommand{\imagewidth}{0.21\columnwidth} 
	\newcommand{\patchwidth}{0.135\columnwidth} 
	\newcommand{\verticaloffset}{1.0cm}
	\newcommand{\addimage}[1]{
        \multirow{3}{*}[\verticaloffset]{\includegraphics[width=\imagewidth]{figures/defocus_map/#1_source_box.jpg}} &
        \multirow{3}{*}[\verticaloffset]{\includegraphics[width=\imagewidth]{figures/defocus_map/#1_tele_box.jpg}} &
        \includegraphics[width=\patchwidth]{figures/defocus_map/#1_source_crop1.jpg} &
        \includegraphics[width=\patchwidth]{figures/defocus_map/#1_tele_crop1.jpg} &
        \includegraphics[width=\patchwidth]{figures/defocus_map/#1_dcsr_sra_crop1.jpg} &
        \includegraphics[width=\patchwidth]{figures/defocus_map/#1_ours_crop1.jpg} \\
        & &
        \includegraphics[width=\patchwidth]{figures/defocus_map/#1_source_crop2.jpg} &
        \includegraphics[width=\patchwidth]{figures/defocus_map/#1_tele_crop2.jpg} &
        \includegraphics[width=\patchwidth]{figures/defocus_map/#1_dcsr_sra_crop2.jpg} &
        \includegraphics[width=\patchwidth]{figures/defocus_map/#1_ours_crop2.jpg} \\
        Full \wide & Full \tele & \wide & \tele & DCSR & Ours 
	}
    \begin{tabular}{cccccc}
        \addimage{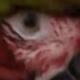}
    \end{tabular}
    \caption{\textbf{When depth-of-field (DoF) is shallower on telephoto (\tele) than wide (\wide),}
    transferring details from \tele to \wide in defocus regions results in significant artifacts. We design our system to exclude defocus regions during fusion and generate results that are robust to lens DoF. By contrast, the result from DCSR~\cite{wang2021dual} shows blurrier details than the input \wide on the parrot and building.
    }
    \label{fig:tele_defocus}
\end{figure}

\paragraph{Adapting to imperfect references}
Existing methods~\cite{zhang2022self,wang2021dual}  treat the entire \tele as a high-resolution reference, resulting in worse fusion quality in regions where \tele is imperfect.
Specifically, two problems can introduce unwanted artifacts:
mismatches in depth-of-field (DoF) and errors in alignment between \wide and \tele.
Due to shallower DoF, out-of-focus pixels on \tele can appear blurrier than on \wide, as shown in~\figref{tele_defocus}.   
We propose an efficient defocus detection algorithm based on the correlation between scene depth and optical flow to exclude defocus areas from fusion.
Based on the defocus map, alignment errors, flow uncertainty, and scene occlusion, we develop an adaptive blending mechanism to generate high-quality and artifact-free super-resolution results.

\paragraph{Minimizing domain gap with real-world inputs}
In RefSR, it is difficult to collect perfectly aligned \wide/\tele ground-truth pairs for training.
As a result, two plausible but inadequate solutions have been explored:
1) Using the reference image \tele as a training target~\cite{wang2021dual,zhang2022self}, which often transfers imperfections from the reference or causes the network to learn the identity mapping.
2) Learning a degradation model to synthesize a low-resolution input from the target image~\cite{trinidad2019multi,zhang2022self}, which introduces a domain gap between training and inference and degrades the super-resolution quality on real-world images.
To avoid learning the identity mapping and minimize the domain gap, we synchronously capture an extra \tele shot from a second smartphone of the same model mounted on a camera rig and use this capture as the reference during training (see Fig.~\ref{fig:camera_rig}).
In this design, the fusion model sees real \wide as input at both the training and inference stages to avoid domain gaps. 
Further, the reference and target are captured from \tele cameras of \emph{different} devices to avoid learning the identity mapping. 

Unlike existing dual-zoom RefSR datasets that either show strong temporal motion between \wide and \tele~\cite{wang2021dual} or are limited to static scenes~\cite{wei2020component}, we collect a large-scale dataset with high-quality \wide/\tele synchronization in dynamic scenes.
Our dataset includes much more diverse captures such as portraits, architectures, landscapes, and challenging scenes like dynamic object motion and night scenes.
We demonstrate that our method performs favorably against state-of-the-art approaches on existing dual-zoom RefSR and our datasets. 

Our contributions are summarized as the following:
\begin{compactitem}[-]
\item An ML-based HZSR system that runs efficiently on mobile devices and is robust to imperfections in real-world images (\secref{algorithm}).
\item A training strategy that uses a dual-phone camera rig to minimize domain gaps and avoids learning a trivial mapping in RefSR (\secref{learning}).
\item A dataset of 150 well-synchronized \wide and \tele shots at high-resolution (12MP), coined as \dataset{Hzsr} dataset, will be released at our project website at\footnote{\url{https://www.wslai.net/publications/fusion_zoom}}
for future research (\secref{results}). 
\end{compactitem}

\section{Related Work}

\begin{figure*}
    \centering
    \includegraphics[width=\textwidth]{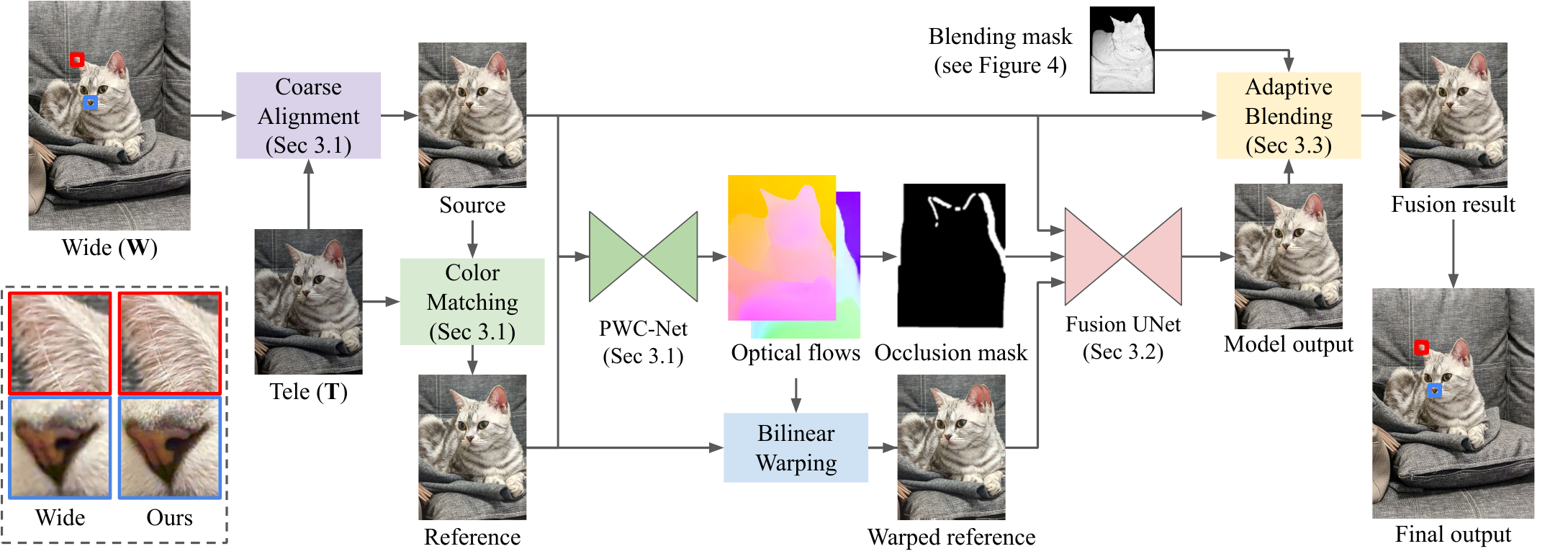}
    \caption{
        \textbf{System overview.} 
        Given concurrently captured \wide and \tele images, we crop \wide to match the FOV of \tele, coarsely align them via feature matching, and adjust the color of \tele to match \wide.
        The cropped \wide and adjusted \tele are referred to as \emph{source} and \emph{reference}, respectively. 
        Then, we estimate dense optical flow to align the reference to source (\secref{alignment}) and generate an occlusion mask.
        Our Fusion UNet takes as input the source, warped reference, and occlusion mask for detail fusion (\secref{fusion}).
        Lastly, we merge the fusion result back to the full \wide image via an adaptive blending (\secref{blending}, \figref{blending}) as the final output.
    }
    \label{fig:system_overview}
\end{figure*}

\paragraph{Learning-based SISR}
Several methods~\cite{dong2014learning,kim2016accurate,lai2017deep,ledig2017photo,zhang2018image,wang2018esrgan,zhang2022efficient,xu2023zero,zhang2019zoom} have shown promising results over the past decade.
However, due to the heavily ill-posed nature, they produce blurry details at large upsampling factors, \eg, 2-5$\times$ required by hybrid zoom on smartphones, or work only for specific domains such as faces~\cite{gu2020image,menon2020pulse,chan2021glean,he2022gcfsr}.

\paragraph{RefSR using Internet images}
RefSR outputs a high-resolution image from a low-resolution input by taking one or multiple~\cite{pesavento2021attention} high-resolution references as additional inputs.
Conventional RefSR methods assume that the references are taken from the internet~\cite{sun2012super} or captured at a different moment, position, or camera model at the same event~\cite{wang2016event,zhang2019image} and focus on improving dense alignment between the source and reference~\cite{zheng2018crossnet,jiang2021robust,xia2022coarse,huang2022task} or robustness to irrelevant references~\cite{shim2020robust,zhang2019image,xie2020feature,lu2021masa,yang2020learning}.
In contrast, we mitigate the alignment challenges by capturing synchronous shots of \wide and \tele to avoid object motion.

\paragraph{RefSR using auxiliary cameras}
Recent RefSR works~\cite{trinidad2019multi,wang2021dual,zhang2022self} capture a reference image of the same scene using an auxiliary camera. 
Since pixel-aligned input and ground-truth image pairs are not available, PixelFusionNet~\cite{trinidad2019multi} learns a degradation model to synthesize a low-resolution input from the high-resolution reference, and use pixel-wise losses such as $\ell_1$ and VGG loss for training. 
Such a model does not generalize well to real-world input images due to the domain gap between the images observed at training and inference times.
On the other hand, SelfDZSR~\cite{zhang2022self}, DCSR~\cite{wang2021dual} and RefVSR~\cite{lee2022reference} treat the reference image as the target for training or fine-tuning.
We observe that such a training setup is prone to degenerate local minimums: the model will often learn the identity mapping and simply copy image content from \tele to the output. 
This results in severe misalignment, color shifting, and DoF mismatches unacceptable for practical photography. 
In this work, we capture an extra \tele shot to mitigate these issues in training.

\paragraph{Efficient mobile RefSR} 
Existing methods typically have large memory footprints due to the use of attention/transformer~\cite{wang2021dual,yang2020learning} or deep architectures~\cite{zhang2022self}. 
They run into out-of-memory (OOM) issues for 12MP input resolution even on an NVIDIA A100 desktop GPU with 40GB RAM and cannot run on mobile devices. 
By contrast, it takes our system only 500ms and 300MB to process 12MP inputs on a mobile GPU. 

Our system design is inspired by the reference-based face deblurring method~\cite{lai2022face}, but the problems we address are fundamentally more challenging.
First, we apply super-resolution to generic images instead of focusing on faces. Our system should be more robust to diverse scenes and handle various of imperfections and mismatches between two cameras.
Second, unlike face deblurring models which can learn from synthetic data, image super-resolution models are more sensitive to the domain gap in training data, and it is more challenging to collect real training data for reference-based SR.
Therefore, our adaptive blending method and dual-phone rig setup are the key components that differentiate our work with~\cite{lai2022face}.

\section{Hybrid Zoom Super-Resolution}
\label{sec:algorithm}

Our goal is to design an efficient system that can run at interactive rates on mobile devices. 
These constraints exclude the use of large models that are slow and memory-intensive. 
The overview of our processing pipeline is shown in~\figref{system_overview}.
When a user zooms to a mid-range zoom (\eg, 3-5$\times$), our system will capture a synchronized image pair when the shutter button is pressed.
We first align \wide and \tele with a global coarse alignment using keypoint matching, followed by a local dense alignment using optical flow (\secref{alignment}).
Then, we adopt a UNet \cite{ronneberger2015u} to fuse the luminance channel of a source image cropped from \wide and a reference warped from \tele (\secref{fusion}).
Lastly, our adaptive blending algorithm (\secref{blending} and~\figref{blending}) takes into account the defocus map, occlusion map, flow uncertainty map, and alignment rejection map to merge the fusion output back to the full-size \wide image.
Overall, our system consists of lightweight modules that make our overall system efficient and effective.

\subsection{Image Alignment}
\label{sec:alignment}
\paragraph{Coarse alignment}
We first crop \wide to match the FOV of \tele, and resample \wide to match the spatial resolution of \tele (4k$\times$3k) using a bicubic resampler.
We then estimate a global 2D translation vector via FAST feature keypoint matching~\cite{rosten2006machine} and adjust the cropped \wide, denoted as $\srcimage$.
We also match \tele's color to \wide by normalizing the mean and variances of RGB colors~\cite{reinhard2001color} to compensate for the photometric differences between \wide and \tele sensors.
The color-adjusted \tele is denoted as the reference image $\refimage$.

\paragraph{Dense alignment}
We use PWC-Net~\cite{sun2018pwc} to estimate dense optical flow between $\srcimage$ and $\refimage$.
Note that the average offset between \wide and \tele is 150 pixels at 12MP resolution, which is much larger than the motion magnitude in most of the optical flow training data~\cite{sun2021autoflow}. The flows estimated from 12MP images are too noisy. Instead, we downsample $\srcimage$ and $\refimage$ to $384 \times 512$ to predict optical flow and upsample flow to the original image resolution to warp $\refimage$ via bilinear resampling, denoting as $\warpedrefimage$. The flow estimated at this scale is more accurate and robust for alignment.

To meet the limited computing budget on mobile devices, we remove the DenseNet structure from the original PWC-Net, which reduces the model size by $50\%$, latency by $56\%$, and peak memory by $63\%$.
While this results in an $8\%$ higher flow end-point-error (EPE) on the Sintel dataset, the flow's visual quality remains similar.
We also generate an occlusion map, $\occlusionmask$, through a forward-backward consistency check~\cite{alvarez2007symmetrical}.

\subsection{Image Fusion}
\label{sec:fusion}
To preserve the color of \wide, we apply fusion in the luminance space only.
We construct a 5-level UNet, which takes as inputs the grayscale $\srcimage$ (denoted as $\srcimageluma$), grayscale $\warpedrefimage$ (denoted as $\warpedrefimageluma$), and the occlusion mask $\occlusionmask$ to generate a grayscale output image $\modeloutputimageluma$.
The grayscale $\modeloutputimageluma$ is merged with the UV channels of $\srcimage$ and converted back to the RGB space as the fusion output image $\modeloutputimage$.
The detailed architecture of the Fusion UNet is provided in the supplementary material.
Since memory footprint is often the bottleneck for on-device processing, a useful design principle for an efficient align-and-merge network is to reduce the feature channels in high-resolution layers. 
Therefore, we design our system with pixel-level image warping instead of feature warping~\cite{reda2022film,trinidad2019multi} and limit the number of encoder channels in Fusion UNet. 

\begin{figure}
    \centering
    \includegraphics[width=\columnwidth]{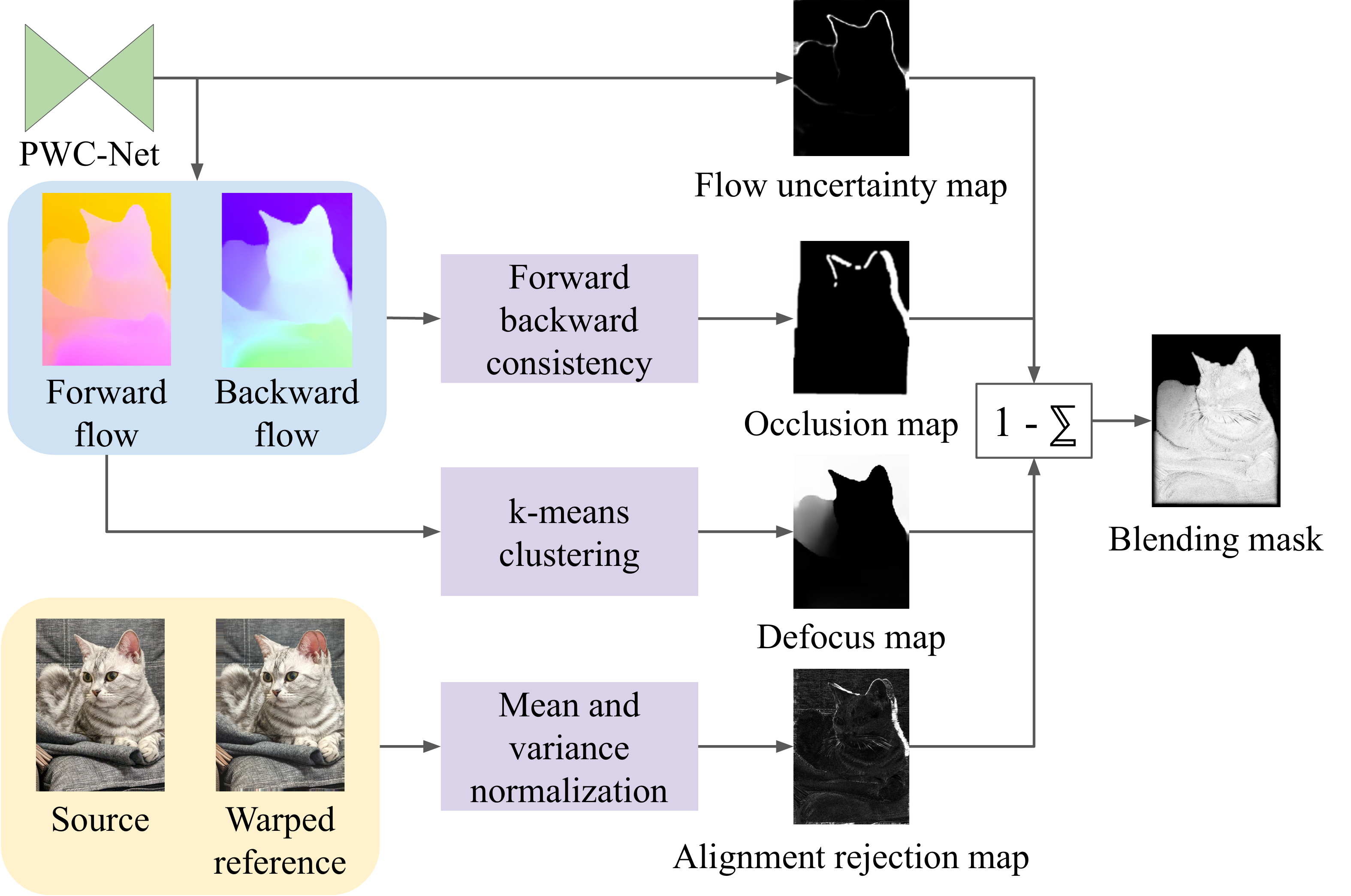}
    \caption{
        \textbf{Adaptive blending.}
        We use alpha masks to make the fusion robust to alignment errors and DoF mismatch (\secref{blending}).
    }
    \label{fig:blending}
\end{figure}

\subsection{Adaptive Blending}
\label{sec:blending}
While ML models are powerful tools for aligning and fusing images, mismatches between \wide and \tele can still result in visible artifacts at output.
Such mismatches include the DoF differences, occluded pixels, and warping artifacts at the alignment stage.
Therefore, we develop a strategy to adaptively blend $\srcimageluma$ and $\modeloutputimageluma$ by utilizing an alpha mask derived from the defocus map, occlusion map, flow uncertainty map, and alignment rejection map, as shown in~\figref{blending}. Our final output is free from objectionable artifacts and robust to imperfections of pixel-level consistency between \wide and \tele.

\paragraph{Narrow DoF on \tele}
We observe that \tele often has a narrower DoF than \wide on mobile phones. 
This is because camera DoF is in proportional to $N / f^2$, where $N$ and $f$ denote the aperture number and focal length, respectively.
The typical focal length ratio between \tele and \wide is $>3\times$ and the aperture number ratio is $<2.5\times$. 
The supplemental material lists the camera specifications from recent flagship phones to justify this observation.
\figref{tele_defocus} shows that 1) the defocused area of \tele is significantly blurrier than that of \wide despite its higher sampling rate, and 2) including defocused details from \tele results in blurry output worse than \wide.
Therefore, we need a defocus map to exclude the defocused pixels from fusion.
Single-image defocus map estimation is an ill-posed problem that requires expensive ML models impractical on mobile devices~\cite{lee2019deep,tang2019defusionnet,zhao2019enhancing,cun2020defocus,xin2021defocus}. 
Instead, we present an efficient algorithm that reuses the optical flow computed at the alignment step.

\begin{figure}
    \centering
    \includegraphics[width=1.0\columnwidth]{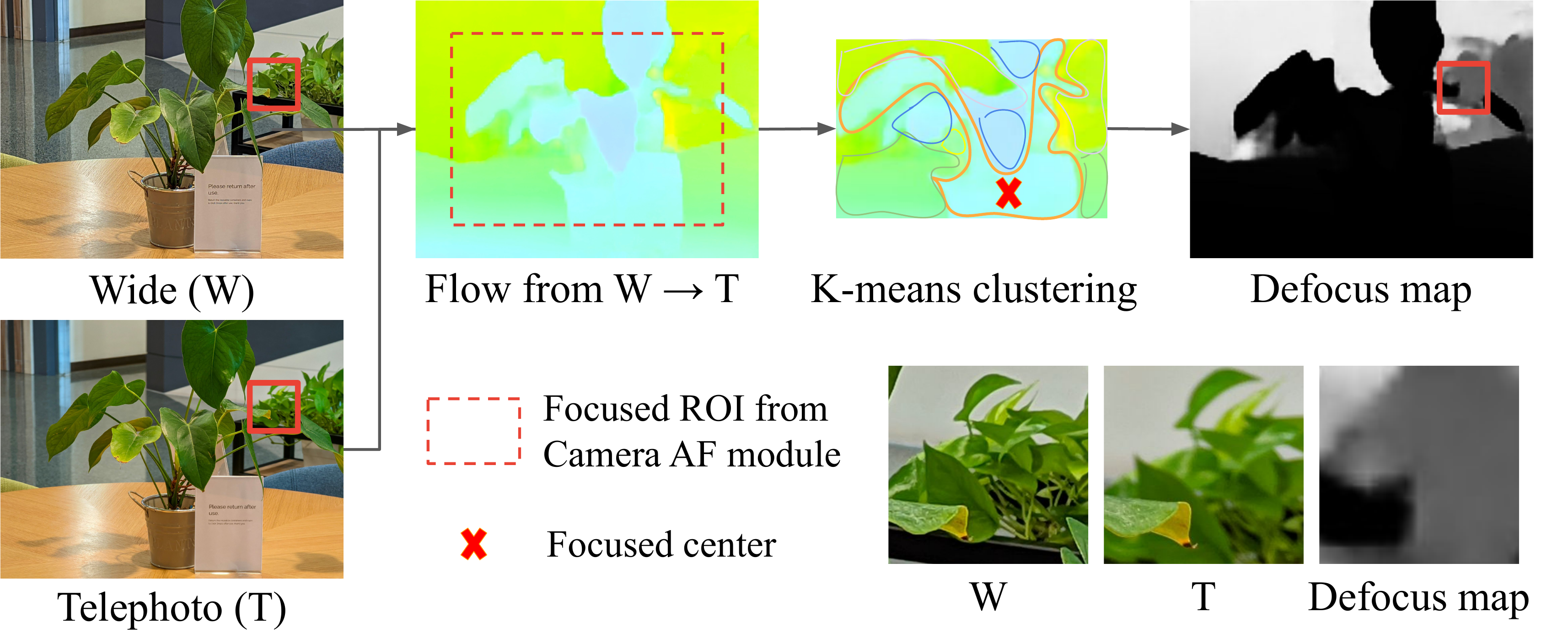}
    \caption{
        \textbf{Efficient defocus map detection} using optical flow at the alignment stage, described in~\secref{blending}. Black/white pixels in the defocus map represent the focused/defocused area.
    }
    \label{fig:defocus_map}
\end{figure}

\paragraph{Defocus map} 
To estimate a defocus map, we need to know 1) where the camera focus is, denoted the focused center, and 2) the relative depth to the focused center for each pixel.
Because \wide and \tele are approximately fronto-parallel, and the optical flow magnitude is proportional to the camera disparity and therefore the scene depth, we propose an algorithm to estimate a defocus map, as illustrated in~\figref{defocus_map}.
First, we acquire the focused region of interest (ROI) from the camera auto-focus module, which indicates a rectangular region on \tele where most pixels are in focus.
Second, based on dual camera stereo, we consider the optical flow as a proxy to depth and assume that the pixels at the same focal plane have similar flow vectors for static scenes~\cite{szeliski2022computer}. 
To find the focused center, we apply the k-mean clustering algorithm on the flow vectors within the focused ROI. 
We then choose the focused center $\textbf{x}_f$ to be the center of the largest cluster.
To estimate the relative depth to  $\textbf{x}_f$, we calculate the $\ell_2$ distance on the flow vectors between each pixel and the focused center, and obtain a defocus map via:
\begin{equation}
    \label{eq:defocusmap}
    \defocusmask(\textbf{x}) \!=\! \sigmoid{\frac{\twonorm{\forwardflow(\textbf{x}) \!-\! \forwardflow(\textbf{x}_f)} \!-\! \gamma}{\sigma_f}},
\end{equation}
where $\forwardflow$ is the optical flow between $\srcimage$ and $\refimage$, $\gamma$ controls the distance threshold to tolerate in-focus regions, and $\sigma_f$ controls the smoothness of the defocus map. Our algorithm is efficient and takes only 5ms on a mobile device.

\paragraph{Occlusion map}
The baseline between \wide and \tele (i.e., the distance between optical centers) makes occluded pixels in \wide invisible to \tele and results in artifacts when warping \tele using optical flow.
To exclude these pixels from fusion, we estimate an occlusion map using the forward-backward flow consistency~\cite{alvarez2007symmetrical}:
\begin{align}
    \occlusionmask(\textbf{x}) = \min( s || \warp(\warp(\textbf{x}; \forwardflow); \backwardflow) - \textbf{x} ||_2, 1),
\end{align}
where $\warp$ is the bilinear warping operator and $\textbf{x}$ is the 2D image coordinate on the source image.
The scaling factor $s$ controls the strength of the occlusion map.
Note that our occlusion map includes both occluded and dis-occluded pixels where the flows are inconsistent, typically near motion or object boundaries.

\paragraph{Flow uncertainty map}
Since dense correspondence is heavily ill-posed, we augment PWC-Net to output a flow uncertainty map~\cite{gast2018lightweight}.
The uncertainty-aware PWC-Net predicts a multivariate Laplacian distribution over flow vectors for each pixel, rather than a simple point estimate. Specifically, it predicts two additional channels that determine to the log-variance of the Laplacian distribution in $x$- and $y$-directions, denoted as $\textbf{Var}_{x}$ and $\textbf{Var}_{y}$, respectively. 
We convert the log-variance into units of pixels through the following equation:
\begin{align}
    &\textbf{S}(\textbf{x}) = \sqrt{\exp(\log(\textbf{Var}_{x}(\textbf{x}))) + \exp(\log(\textbf{Var}_{y}(\textbf{x})))}, \\
    &\flowconfmask(\textbf{x}) = \min( \textbf{S}(\textbf{x}), s_{\text{max}}) / s_{\text{max}}.
\end{align}
As shown in~\figref{blending}, the flow uncertainty map often highlights the object boundary or texture-less regions. We use $s_{\text{max}}=8$ at the flow prediction resolution.

\paragraph{Alignment rejection map}
We estimate an alignment rejection map to exclude the erroneous alignment based on the similarity between the source and aligned reference patches~\cite{hasinoff2016burst,wronski2019handheld},
First, to match the optical resolution between \wide and \tele, we use bilinear resizing to downsample and upsample the warped reference frame $\warpedrefimageluma$ based on the focal length ratio between \wide and \tele, denoted by $\warpedrefimagelumaresampled$. 
Then, for each pixel with its local patch $\srcimagelumaresampledpatch$ (on $\srcimageluma$) and $\warpedrefimagelumaresampledpatch$ (on $\warpedrefimagelumaresampled$), we subtract patch means and calculate the normalized patch difference $P_\delta = (\srcimagelumaresampledpatch - \srcimagelumaresampledpatchmean) - (\warpedrefimagelumaresampledpatch - \warpedrefimagelumaresampledpatchmean)$.
The alignment rejection map on each patch is calculated by:
\begin{align}
    \rejectionmask (\textbf{x}) = 1 - \exp{\left(-\twonorm{P_\delta(\textbf{x})} / \left(\srcimagelumaresampledpatchvariance(\textbf{x}) + \epsilon_0 \right)\right)},
\end{align}
where $\srcimagelumaresampledpatchvariance$ is the variance of $\srcimagelumaresampledpatch$, $\epsilon_0$ is used to tolerate minor diff between source and reference. We set patch size to $16$ and stride size to $8$ in all our experiments.

\paragraph{Final blending}
We generate the blending mask as:
\begin{align}
\!\blendingmask\!=\!\newmax{\textbf{1}\!-\!\occlusionmask\!-\!\defocusmask\!-\!\flowconfmask\!-\!\rejectionmask,\!\textbf{0}}.\!
\label{eq:blending}
\end{align}
Note that $\defocusmask$, $\occlusionmask$ and $\flowconfmask$ are generated at the flow inference size, and $\rejectionmask$ is $8\times$ smaller than $\srcimage$.
We upscale these masks using a bilinear upsampling to the size of $\srcimage$ for blending.
For pixels outside the FOV of \tele, we retain the intensities of \wide and apply a Gaussian smoothing on the boundary of $\blendingmask$ to avoid abrupt transitions between the fusion and non-fusion regions. 
The final output image is generated via an alpha blending and ``uncropping'' back to the full \wide image:
\begin{align}
\!\finaloutputiamge\!=\!\uncrop{\blendingmask\!\odot\!\modeloutputimage\!+\! (\textbf{1}\!-\!\blendingmask)\!\odot\!\srcimage,\!\wide},\!
\end{align}
where $\odot$ is the Hadamard product.

\section{Learning from Dual Camera Rig Captures}
\label{sec:learning}

\begin{figure}
    \centering
    \footnotesize
    \includegraphics[width=1.0\columnwidth]{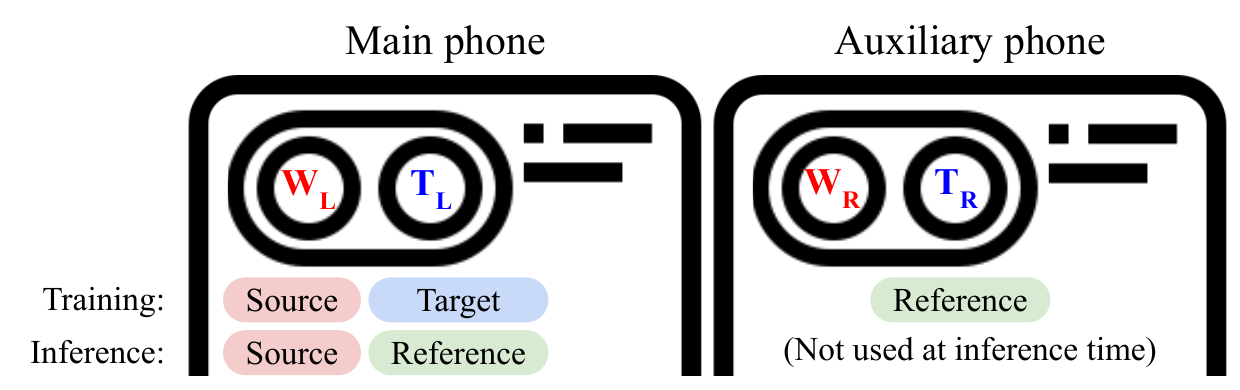}
    \caption{
        \textbf{Dual-phone rig setup.} We collect synchronous captures from two smartphones on a rig and use $\wide_L$, $\tele_L$, and $\tele_R$ as source, target, and reference images. The training setup ensures the camera sensors are consistent between the test and training stages to eliminate the domain gap.  
    }
    \label{fig:camera_rig}
\end{figure}

Techniques that synthesize degraded inputs for training~\cite{trinidad2019multi,zhang2022self,wang2021dual} suffer from a domain gap between synthetic and real images.
To reduce the gap, we train our alignment and fusion models on real-world images where the source, reference, and ground-truth images are all captured by mobile phone cameras.

\paragraph{Dual camera rig}
We design a dual-phone rig to mount two smartphones side-by-side, as illustrated in~\figref{camera_rig}. 
The rig is 3D-printed and designed to fix the main and auxiliary devices in a fronto-parallel position and an identical vertical level. 
We use a camera app that synchronizes the capture time between the main and auxiliary phones through WiFi~\cite{ansari2019wireless}.
In~\figref{camera_rig}, we denote the cameras on the left phone as $\wide_L$ and $\tele_L$, and the cameras on the right phone as $\wide_R$ and $\tele_R$.

In the training time, we take $\wide_L$ and $\tele_R$ as the source and reference pairs (\ie, inputs to the model) and $\tele_L$ as the target image (\ie, ground truth of the model output).
We use PWC-Net to align $\wide_L$ and $\tele_R$ to $\tele_L$, so that the source, reference, and target images are all aligned to the same camera viewpoint.
As both the source and reference images are warped, we define an availability mask $\availabilitymask = \textbf{1} - \trainingocclusionmask$, where $\trainingocclusionmask$ denotes the union of the occlusion masks from $\wide_L \rightarrow \tele_L$ flow and $\tele_R \rightarrow \tele_L$ flow, as the loss is inapplicable to occluded pixels and should be excluded.
Note that we select $\tele_L$ instead of $\tele_R$ as the target image to minimize the warping distance between the source and target.
If we select $\tele_R$ as the target, both $\wide_L$ and $\tele_L$ have to be warped from the left smartphone position to align with $\tele_R$ on the right smartphone, which will reduce the number of valid pixels for training.
More details about our training setup are provided in the supplementary material.
In total, we collect 8,322 triplets to train our Fusion UNet.

At inference time, we only need \wide and \tele from one smartphone, (\ie, $\wide_L$ and $\tele_L$), and \tele is warped to align with \wide for fusion.
The only difference between training and testing lies in the image alignment: we align all the images to $\wide_L$ at inference time but align to $\tele_L$ at training time to minimize warping errors.
Note that the warped $\wide_L$ and warped $\tele_R$ in the training stage are not exact but close enough to mimic the real source and reference images at test time; they are all real images from the corresponding camera sensors.

\paragraph{Fusion UNet training}
We denote the target image $\tele_L$ as $\targetimage$, and train our Fusion UNet with the following losses. 

\subparagraph{VGG loss}
The perceptual loss~\cite{johnson2016perceptual} between $\modeloutputimage$ and $\targetimage$, which is commonly used in image restoration:
\begin{equation}
\label{eq:vgg_loss}
    \vggloss = \onenorm{\availabilitymask \odot (VGG(\modeloutputimage) - VGG(\targetimage))}.
\end{equation}
Note that the availability mask $\availabilitymask$ is resized to the resolution of the corresponding VGG features.

\subparagraph{Contextual loss}: While we pre-align the source and target images, misalignments still exist and degrade the model performance by generating blurry predictions or warping artifacts.
Therefore, we adopt the contextual loss~\cite{mechrez2018contextual} to learn better on mis-aligned training data:
\begin{equation}
\label{eq:contextual_loss}
    \!\contextualloss\!=\!CX\!(\availabilitymask \odot VGG(\!\modeloutputimage\!), \availabilitymask \odot VGG(\!\targetimage\!)\!),\!
\end{equation}
where $CX$ is the contextual similarity~\cite{mechrez2018contextual} between the VGG features of $\modeloutputimage$ and $\targetimage$.

\subparagraph{Brightness-consistency loss}: To preserve the low-frequency brightness tone on \wide and avoid tonal shift, we apply a brightness-consistency loss~\cite{lai2022face}:
\begin{equation}
\label{eq:color_loss}
    \colorloss = \onenorm{\mathbb{G}(\modeloutputimageluma, \sigma) -\mathbb{G}(\srcimageluma, \sigma)},
\end{equation}
where $\mathbb{G}$ denotes Gaussian filter with a standard deviation $\sigma=10$ in this work. 
Note that the brightness-consistency loss is applied to the whole image to encourage the model to learn the identity mapping over the occluded regions.

The final loss $\finalloss$ is:
\begin{equation}
\label{eq:final_loss}
    \finalloss = \vgglossweight \vggloss + \contextuallossweight \contextualloss + \colorlossweight \colorloss
\end{equation}
where we set $\vgglossweight = 1$, $\contextuallossweight = 0.05$, and $\colorlossweight = 1$. 
Note that $\vggloss$ is effective for aligned pixels, while $\contextualloss$ is more suitable for misaligned content. Our model requires both losses to achieve better fusion quality, while the weight on VGG loss is much higher than contextual loss.

\paragraph{PWC-Net training.}
The PWC-Net is pre-trained on the AutoFlow dataset~\cite{sun2021autoflow}.
However, there is a domain gap between the AutoFlow training data and the images from mobile phones.
Therefore, we use the $\srcimage$ and $\refimage$ as input images and generate ``pseudo'' ground-truth flow with RAFT~\cite{teed2020raft,sun2022disentangling} to further fine-tune the PWC-Net.
The PWC-Net can then be adapted and generalized well to align our source and reference images. Please see the supplementary materials for the effect of fine-tuning PWC-Net.

\section{Experimental Results}
\label{sec:results}

We evaluate our system on our \dataset{Hzsr} dataset, compare against recent RefSR methods, analyze system performance, conduct ablation studies on key components, and discuss the limitations in this section.
More high-resolution visual comparisons are available in the supplementary materials and our project website.

\subsection{Hybrid Zoom SR (\dataset{\textbf{Hzsr}}) dataset}
We use a smartphone with a \wide and \tele, which is commonly available among flagship smartphones. 
When the zoom level exceeds the focal length ratio between \tele and \wide, \ie, $5\times$, the hybrid zoom system will switch from \wide to \tele. 
Just before this zoom ratio, the \wide is upsampled to account for the difference in sensor resolution. 
We collect $25,041$ pairs of \wide and \tele image pairs with zoom ranges varying from $2\times$ to $5\times$ for validating the proposed system.
Among them, we select $150$ representative images that cover a wide variety of real-world scenes, including landscapes, street views, portraits, animals, and low-light images, named as Hybrid Zoom SR (\dataset{Hzsr}) dataset.
The 150 images will be publicly released on our project website.

\begin{figure*}
    \centering
    \footnotesize
    \renewcommand{\tabcolsep}{1pt} 
	\renewcommand{\arraystretch}{1} 
 	\renewcommand{\imagewidth}{0.272\linewidth}
	\newcommand{\patchwidth}{0.1\linewidth}
	\newcommand{\verticalshift}{1.6cm}
    \newcommand{\addimage}[1]{
        \multirow{2}{*}[\verticalshift]{\includegraphics[width=\imagewidth]{figures/HZSR/#1_LR_box.jpg}} &
        \multirow{2}{*}[\verticalshift]{\includegraphics[width=\imagewidth]{figures/HZSR/#1_tele_box.jpg}} &
        \includegraphics[width=\patchwidth]{figures/HZSR/#1_LR_crop0.jpg} &
        \includegraphics[width=\patchwidth]{figures/HZSR/#1_tele_crop0.jpg} &
        \includegraphics[width=\patchwidth]{figures/HZSR/#1_DCSR_SRA_crop0.jpg} &
        \includegraphics[width=\patchwidth]{figures/HZSR/#1_FusionZoom_crop0.jpg} \\
        &
        &
        \includegraphics[width=\patchwidth]{figures/HZSR/#1_LR_crop1.jpg} &
        \includegraphics[width=\patchwidth]{figures/HZSR/#1_tele_crop1.jpg} &
        \includegraphics[width=\patchwidth]{figures/HZSR/#1_DCSR_SRA_crop1.jpg} &
        \includegraphics[width=\patchwidth]{figures/HZSR/#1_FusionZoom_crop1.jpg} \\
        Full \wide & Full \tele & \wide & \tele & DCSR & Ours \\
	}
    \begin{tabular}{cccccc}
        \addimage{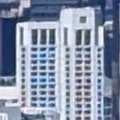}
        \addimage{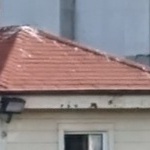}
        \addimage{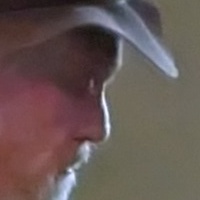}
        \addimage{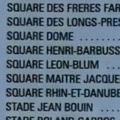}
        \addimage{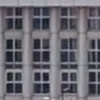}
    \end{tabular}
    \caption{\textbf{Visual comparisons on our \dataset{HZSR} dataset.} Our method recovers more fine details and textures (e.g., the building facades, more legible letters, and facial details) than DCSR~\cite{wang2021dual}.}
    \label{fig:hzsr}
\end{figure*}

\begin{figure*}
    \centering
    \footnotesize
    \renewcommand{\tabcolsep}{1pt} 
	\renewcommand{\arraystretch}{1} 
    \renewcommand{\imagewidth}{0.196\linewidth}
	\newcommand{\patchwidth}{0.12\linewidth}
	\newcommand{\verticalshift}{1.95cm}
    \newcommand{\addimage}[1]{
        \multirow{2}{*}[\verticalshift]{\includegraphics[width=\imagewidth]{figures/HZSR/#1_LR_box.jpg}} &
        \multirow{2}{*}[\verticalshift]{\includegraphics[width=\imagewidth]{figures/HZSR/#1_tele_box.jpg}} &
        \includegraphics[width=\patchwidth]{figures/HZSR/#1_LR_crop1.jpg} &
        \includegraphics[width=\patchwidth]{figures/HZSR/#1_tele_crop1.jpg} &
        \includegraphics[width=\patchwidth]{figures/HZSR/#1_DCSR_SRA_crop1.jpg} &
        \includegraphics[width=\patchwidth]{figures/HZSR/#1_FusionZoom_crop1.jpg} \\
        & & \wide & \tele & DCSR & Ours \\
        &
        &
        \includegraphics[width=\patchwidth,cfbox=black 0.3pt 0pt]{figures/HZSR/#1_FusionZoom_occlusion_mask_crop1.jpg} &
        \includegraphics[width=\patchwidth,cfbox=black 0.3pt 0pt]{figures/HZSR/#1_FusionZoom_defocus_map_crop1.jpg} &
        \includegraphics[width=\patchwidth,cfbox=black 0.3pt 0pt]{figures/HZSR/#1_FusionZoom_rejection_map_crop1.jpg} &
        \includegraphics[width=\patchwidth,cfbox=black 0.3pt 0pt]{figures/HZSR/#1_FusionZoom_blend_mask_crop1.jpg} \\
        Full \wide & Full \tele & $\occlusionmask$ & $\defocusmask$  & $\rejectionmask$ & $\blendingmask$ \\
	}
    \begin{tabular}{cccccc}
        \addimage{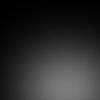}
        \addimage{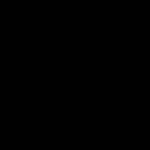}
        \addimage{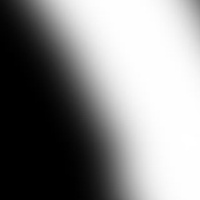}
        \addimage{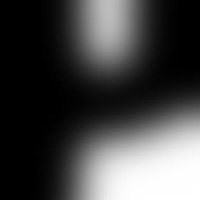}
    \end{tabular}
    \caption{\textbf{Visual comparisons on our \dataset{HZSR} dataset.} We visualize our occlusion mask $\occlusionmask$, defocus mask $\defocusmask$,  alignment rejection mask $\rejectionmask$, and blending mask $\blendingmask$ on each example. Note that brighter pixels on $\blendingmask$ indicate the pixels to be included in the fusion output (see \eqnref{blending}). On the top two examples, \tele is completely out of focus. DCSR's results are blurrier than \wide, while our method can preserve the same sharpness as \wide. On the bottom two examples, DCSR makes the pixels around occlusion regions blurrier than \wide. Our results maintain the same amount of details as \wide.
    }
    \label{fig:hzsr2}
\end{figure*}

We show a few landscape and mid-zoom range shots in~\figref{hzsr}, which are the common use cases of hybrid zoom. Our method is able to transfer the details from \tele to recover the facades on buildings and make letters more legible.
\figref{hzsr2} highlight the shots with occlusion and defocus blur on \tele. DCSR~\cite{wang2021dual} often transfers unwanted blur to the output image, resulting in quality drop compared to the input \wide image. In contrast, our method preserves the sharpness and details of \wide via the adaptive blending. 
Note that we do not attempt to hallucinate details in defocus and occlusion areas. Instead, our system robustly falls back to \wide's pixels in these error-prone areas. 

Note that except our method and DCSR~\cite{wang2021dual}, all other methods failed to process 12MP inputs due to out-of-memory errors on A100 GPUs with 40GB memory. 

\subsection{Comparisons with RefSR methods}
We compare our method with SRNTT~\cite{zhang2019image}, TTSR \cite{yang2020learning}, MASA~\cite{lu2021masa}, C2-Matching~\cite{jiang2021robust},
AMSA~\cite{xia2022coarse}, DCSR~\cite{wang2021dual}, and SelfDZSR~\cite{zhang2022self}.
We use the pre-trained models from the authors' websites without retraining, as not all the implementations support retraining with 12MP inputs.

\begin{figure}
    \centering
    \small
    \renewcommand{\tabcolsep}{0.5pt} 
	\renewcommand{\arraystretch}{0.8} 
	\renewcommand{\imagewidth}{0.36\columnwidth}
	\newcommand{\patchwidth}{0.17\columnwidth}
	\newcommand{\verticalshift}{0.7cm}
    \newcommand{\addimage}[8]{
        \multirow{3}{*}[\verticalshift]{\includegraphics[width=\imagewidth]{figures/CameraFusion/#1_LR_box.jpg}} &
        \includegraphics[width=\patchwidth]{figures/CameraFusion/#1_LR_crop.jpg} &
        \includegraphics[width=\patchwidth]{figures/CameraFusion/#1_Ref_crop.jpg} &
        \includegraphics[width=\patchwidth]{figures/CameraFusion/#1_HR_crop.jpg} &
        \includegraphics[width=\patchwidth]{figures/CameraFusion/#1_SRNTT_crop.jpg} &
        \includegraphics[width=\patchwidth]{figures/CameraFusion/#1_TTSR_crop.jpg}\\
        & Input \wide & Input \tele & \wide & SRNTT & TTSR \\
        & (source) & (reference) & (GT) & #2 & #3 \\
        &
        \includegraphics[width=\patchwidth]{figures/CameraFusion/#1_MASA_crop.jpg} &
        \includegraphics[width=\patchwidth]{figures/CameraFusion/#1_C2_Matching_crop.jpg} &
        \includegraphics[width=\patchwidth]{figures/CameraFusion/#1_DCSR_SRA_crop.jpg} &
        \includegraphics[width=\patchwidth]{figures/CameraFusion/#1_SelfDZSR_crop.jpg} &
        \includegraphics[width=\patchwidth]{figures/CameraFusion/#1_FusionZoom_crop.jpg} \\
        Full \wide & MASA & C2 & DCSR & SelfDZSR & Ours \\
        PSNR / SSIM & #4 & #5 & #6 & #7 & #8 \\
	}
	\resizebox{\columnwidth}{!}{
    \begin{tabular}{cccccc}
        \addimage{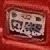}{32.4 / 0.87}{\textbf{36.1} / \textbf{0.91}}{26.6 / 0.66}{32.2 / 0.84}{32.6 / 0.83}{31.9 / 0.85}{34.2 / 0.89}
    \end{tabular}
    }
    \caption{\textbf{Comparisons on \dataset{CameraFusion} dataset.} Our results are visually comparable to SelfDZSR~\cite{zhang2022self} and recover more legible letters than others. TTSR~\cite{yang2020learning} achieves the highest PSNR/SSIM as it generates an output closer to the ground-truth (\wide), while our method recovers details to match the reference (\tele).
    }
    \label{fig:compare_camera_fusion}
\end{figure}

\paragraph{\dataset{CameraFusion} dataset~\cite{wang2021dual}.}
This dataset contains 146 pairs (132 for training and 14 for testing) of \wide and \tele images collected from mobile phones. 
Both \wide and \tele are downsampled $2\times$ to 3MP resolution as inputs, while the original 12MP \wide images are used as the ground-truth during evaluation.
Because of this, \dataset{CameraFusion} dataset can be considered as a synthetic dataset for $2\times$ SR evaluation.
In~\figref{compare_camera_fusion}, our method outputs the most legible letters among the methods. 
Other RefSR works~\cite{wang2021dual,zhang2022self} observe that optimizing with $\ell_1$ or $\ell_2$ loss results in the best reference-based metrics but worse visual quality. We also re-train our model with $\ell_1$ loss on the training set of \dataset{CameraFusion} dataset and report the results in~\tabref{evaluation_table}.
Note that our results are not favored by the evaluation setup of \dataset{CameraFusion}, as our method aims to match the detail level of the reference.
The reference may contain more details than the ground-truth, \eg, in~\figref{compare_camera_fusion} the letters in the input \tele are more legible than the ground-truth \wide.
As a result, our method is more visually pleasing but has a lower PSNR or SSIM in this dataset.

\begin{figure}
    \centering
    \small
    \renewcommand{\tabcolsep}{0.5pt} 
	\renewcommand{\arraystretch}{0.8} 
	\renewcommand{\imagewidth}{0.36\columnwidth}
	\newcommand{\patchwidth}{0.17\columnwidth}
	\newcommand{\verticalshift}{0.7cm}
    \newcommand{\addimage}[9]{
        \multirow{3}{*}[\verticalshift]{\includegraphics[width=\imagewidth]{figures/Nikon/#1_LR_box.jpg}} &
        \includegraphics[width=\patchwidth]{figures/Nikon/#1_LR_crop.jpg} &
        \includegraphics[width=\patchwidth]{figures/Nikon/#1_HR_crop.jpg} &
        \includegraphics[width=\patchwidth]{figures/Nikon/#1_SRNTT_crop.jpg} &
        \includegraphics[width=\patchwidth]{figures/Nikon/#1_TTSR_crop.jpg} &
        \includegraphics[width=\patchwidth]{figures/Nikon/#1_MASA_crop.jpg}\\
        & \wide & \tele & SRNTT & TTSR & MASA \\
        & (source) & (reference) & #2 & #3 & #4 \\
        &
        \includegraphics[width=\patchwidth]{figures/Nikon/#1_C2_Matching_crop.jpg} &
        \includegraphics[width=\patchwidth]{figures/Nikon/#1_AMSA_crop.jpg} &
        \includegraphics[width=\patchwidth]{figures/Nikon/#1_DCSR_SRA_crop.jpg} &
        \includegraphics[width=\patchwidth]{figures/Nikon/#1_SelfDZSR_crop.jpg} &
        \includegraphics[width=\patchwidth]{figures/Nikon/#1_FusionZoom_crop.jpg} \\
        Full \wide & C2 & AMSA & DCSR & SelfDZSR & Ours \\
        PSNR / SSIM & #5 & #6 & #7 & #8 & #9 \\
	}
	\resizebox{\columnwidth}{!}{
    \begin{tabular}{cccccc}
        \addimage{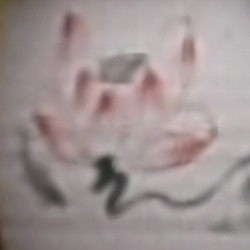}{28.93 / 0.82}{28.69 / 0.81}{28.67 / 0.81}{28.66 / 0.81}{28.70 / 0.82}{27.91 / 0.767}{28.95 / 0.82}{\textbf{37.89} / \textbf{0.98}}
    \end{tabular}
    }
    \caption{\textbf{Comparisons on \dataset{DRealSR} dataset.} Our method achieves the best visual quality and appear closer to the reference when compared to other methods.
    }
    \label{fig:compare_nikon}
\end{figure}

\begin{figure}
    \centering
    \includegraphics[width=\columnwidth]{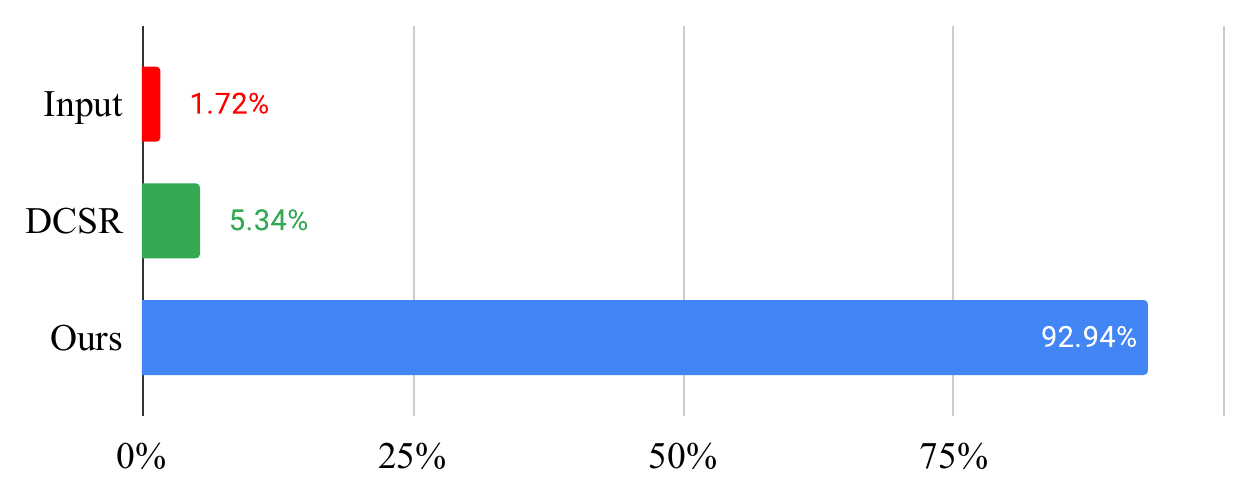}
    \caption{\textbf{User study on \dataset{Hzsr} dataset.} Our results are favored in $92.94\%$ of the comparisons.}
    \label{fig:user_study}
\end{figure}

\begin{table}
    \centering
    \scriptsize
    \caption{\textbf{Quantitative evaluation on the \dataset{CameraFusion} and \dataset{DRealSR} datasets.} 
    Top: methods trained with $\ell_1$ or $\ell_2$ loss. \emph{Ours-$\ell_1$} is re-trained on each dataset.
    Bottom: methods trained with their own defined losses. \emph{Ours} model is trained on the \dataset{Hzsr} dataset using the loss in \eqnref{final_loss}. 
    The best metrics for \dataset{CameraFusion} do not always correlate with the best visual result as observed in~\figref{compare_camera_fusion} and~\cite{wang2021dual,zhang2022self}. Note that DCSR~\cite{wang2021dual} does not release their $\ell_1$ model, so we only report the PSNR/SSIM from their paper.
    }
    \begin{tabular}{r|ccc|ccc}
        \toprule
        \multirow{2}{*}{Method} & \multicolumn{3}{c|}{\dataset{CameraFusion}} & \multicolumn{3}{c}{\dataset{DRealSR}} \\
         & PSNR $\uparrow$ & SSIM $\uparrow$ & LPIPS$\downarrow$ & PSNR$\uparrow$ & SSIM$\uparrow$ & LPIPS$\downarrow$\\
        \midrule
        SRNTT-$\ell_2$        & 33.36 & 0.909 & 0.131\cellcolor{cellred} & 27.30 & 0.839 & 0.397\cellcolor{cellyellow} \\
        TTSR-$\ell_1$       & 36.28 \cellcolor{cellorange}& 0.928\cellcolor{cellorange} & 0.140\cellcolor{cellyellow} & 25.83 & 0.827 & 0.411 \\
        MASA-$\ell_1$             & 31.85 & 0.887 & 0.137\cellcolor{cellorange} & 27.27 & 0.837 & 0.415 \\ 
        C2-Matching-$\ell_1$ & 32.60 & 0.899 & 0.142 & 27.19 & 0.840\cellcolor{cellyellow} & 0.412 \\ 
        AMSA-$\ell_1$          & 32.99 & 0.888 & 0.159 & 28.04\cellcolor{cellyellow} & 0.822 & 0.411 \\
        DCSR-$\ell_1$        & 36.98\cellcolor{cellred} & 0.933\cellcolor{cellred} & n/a  & 27.73 & 0.827 & n/a  \\   
        SelfDZSR-$\ell_1$    & 30.89 & 0.868 & 0.255 & 28.93\cellcolor{cellorange} & 0.857\cellcolor{cellred} & 0.328\cellcolor{cellorange} \\
        Ours-$\ell_1$                               & 34.91\cellcolor{cellyellow} & 0.916\cellcolor{cellyellow} & 0.170  & 31.07\cellcolor{cellred} & 0.852\cellcolor{cellorange} & 0.131\cellcolor{cellred} \\
        \midrule
        SRNTT                 & 31.61 & 0.891 & 0.116\cellcolor{cellorange} & 27.31 & 0.824\cellcolor{cellyellow} & 0.366\cellcolor{cellyellow} \\
        TTSR                & 35.48\cellcolor{cellred} & 0.915\cellcolor{cellred} & 0.162 & 25.31 & 0.772 & 0.388 \\
        MASA                      & 28.05 & 0.759 & 0.255 & 27.32 & 0.764 & 0.381 \\ 
        C2-Matching          & 31.86 & 0.858 & 0.138 & 26.79 & 0.814 & 0.401 \\ 
        AMSA                   & 31.57 & 0.885 & 0.121\cellcolor{cellyellow} & 28.03\cellcolor{cellyellow} & 0.821 & 0.388 \\ 
        DCSR                    & 34.41\cellcolor{cellorange} & 0.904\cellcolor{cellyellow} & 0.106\cellcolor{cellred} & 27.69 & 0.823 & 0.373 \\  
        SelfDZSR               & 30.97 & 0.870 & 0.134 & 28.67\cellcolor{cellorange} & 0.836\cellcolor{cellorange} & 0.249\cellcolor{cellorange} \\
        Ours                                        & 34.09\cellcolor{cellyellow} & 0.907\cellcolor{cellorange} & 0.152 & 31.20\cellcolor{cellred} & 0.875\cellcolor{cellred} & 0.133\cellcolor{cellred} \\
        \bottomrule
    \end{tabular}
    \label{tab:evaluation_table}
\end{table}

\begin{table}
    \centering
    \scriptsize
    \caption{\textbf{Latency comparison}. We measure the inference latency (in milliseconds, or ms) on a Google Cloud platform virtual machine with an Nvidia A100 GPU (40 GB RAM). Most methods hit out-of-memory (OOM) errors when the input image size is larger than $512 \times 512$, while our model can process 12MP resolution inputs within 3ms.
    }
    \begin{tabular}{r|ccccc}
        \toprule
        \multirow{2}{*}{{Method}{Size}}  & $256\times$ & $512\times$ & $1024\times$ & $2016\times$ & $4032\times$ \\
          & $256$ & $512$ & $1024$ & $1512$ & $3024$ \\
        \midrule
        SRNTT       & 2 mins    & 20 mins   & 1 day & OOM   & OOM \\
        TTSR       & 2,665      & OOM       & OOM   & OOM   & OOM \\
        MASA            & 371       & OOM       & OOM   & OOM   & OOM \\
        C2-Matching & 373       & OOM       & OOM   & OOM   & OOM \\
        AMSA         & 6,024     & OOM       & OOM   & OOM   & OOM \\
        DCSR          & 52        & OOM       & OOM   & OOM   & OOM \\
        SelfDZSR      & 35        & 121       & 724   & 3,679 & OOM \\
        Ours                               & \textbf{2.72}     & \textbf{2.82}      & \textbf{2.86}  & \textbf{2.87} & \textbf{2.95} \\
        \bottomrule
    \end{tabular}
    \label{tab:compare_latency}
\end{table}

\paragraph{\dataset{DRealSR} dataset~\cite{wei2020component}.} 
This dataset includes 163 pairs of images captured from long and short focal lengths of a DSLR with a $4\times$ zoom lens.
The images are nearly disparity-free, but the content does not have dynamic subject motion.
Following the strategy in SelfDZSR~\cite{zhang2022self}, we consider the short and long focal-length images as the input and reference, respectively.
The reference is warped by PWC-Net~\cite{sun2018pwc} to align with the input image and used as ground-truth for evaluation~\cite{zhang2022self}. 
Note that such a ground-truth image still has misalignment with the input and may contain warping artifacts that affect PSNR and SSIM metrics.
\tabref{evaluation_table} shows that our method outperforms existing approaches under such an evaluation setup. 
\figref{compare_nikon} shows that we effectively transfer the details from the reference to output, while state-of-the-art approaches often generate blurry outputs.

\paragraph{User study.}
As our \dataset{Hzsr} dataset does not have any ground-truth for quantitative evaluation, we conduct a user study to evaluate the subject preference on the results.
We design a blind user study, where users do not know which method the image is generated from.
Each question shows an image from: the input \wide, output from DCSR~\cite{wang2021dual}, and output from our system, and asks users to choose the one with the best detail fidelity, such as sharpness, clarity, legibility, and textures.
We randomly select 20 images from the \dataset{Hzsr} dataset in each user session.
In total, we collect feedback from 27 users (540 image comparisons). 
Overall, our results are favored in 92.9$\%$ of images, where DCSR and the input \wide are chosen in 1.7$\%$ and 5.4$\%$ of images, respectively (see \figref{user_study}).

\paragraph{Performance on workstation.}
We use a Google cloud platform virtual machine with 12 cores CPU and an Nvidia A100 GPU (40 GB RAM) to test all the methods with input image sizes ranging from $256 \times 256$ to 12MP (4k $\times$ 3k).
As shown in~\tabref{compare_latency}, TTSR~\cite{yang2020learning}, MASA~\cite{lu2021masa}, C2-Matching~\cite{jiang2021robust}, AMSA~\cite{xia2022coarse}, and DCSR~\cite{wang2021dual} all hit out-of-memory errors when the input size is larger than $512 \times 512$.
None of the existing models can process 12MP images directly, while our model can process 12MP input images within 3ms.
Note that DCSR~\cite{wang2021dual} uses a patch-based inference to process high-resolution images.
We adopt the same patch-based inference to generate the results of other compared methods on the \dataset{CameraFusion} and \dataset{Hzsr} datasets.

\begin{table}
    \centering
    \scriptsize
    \caption{\textbf{On-device latency breakdown} (in milliseconds) on a Google Pixel 7 Pro phone with a Mali-G710 GPU.}
    \begin{tabular}{l|c|l|c}
        \toprule
        Stage & Latency         & Stage & Latency \\
        \midrule
        Color matching          & 10  &  Defocus map             & 2    \\
        Coarse alignment        & 44  &  Alignment rejection map & 131  \\
        Optical flow estimation & 65  &  Fusion UNet             & 240  \\
        Bilinear warping        & 23  &  Adaptive blending       & 5    \\
        Occlusion map           & 11  &  \textbf{Total}          & \textbf{521}  \\
        \bottomrule
    \end{tabular}
    \label{tab:labency_breakdown}
\end{table}

\paragraph{Performance on device.}
We implement and benchmark our system on Google Pixel 7 Pro and show the latency breakdown in~\tabref{labency_breakdown}.
The total latency overhead is 521ms. 
Peak memory usage occurs during the Fusion UNet inference stage, which takes an extra 300MB compared to regular single-camera use cases.

\begin{figure}
    \centering
    \footnotesize
    \renewcommand{\tabcolsep}{1pt} 
	\renewcommand{\arraystretch}{0.9} 
	\renewcommand{\imagewidth}{0.39\columnwidth} 
	\newcommand{\patchwidth}{0.142\columnwidth} 
	\newcommand{\verticaloffset}{1.03cm} 
	\newcommand{\addimage}[1]{
        \multirow{2}{*}[\verticaloffset]{\includegraphics[width=\imagewidth]{figures/defocus_map/#1_source_box_with_defocus_map.jpg}} &
        \includegraphics[width=\patchwidth]{figures/defocus_map/#1_source_crop1.jpg} &
        \includegraphics[width=\patchwidth]{figures/defocus_map/#1_warped_reference_crop1.jpg} &
        \includegraphics[width=\patchwidth]{figures/defocus_map/#1_fusion_no_defocus_map_crop1.jpg} &
        \includegraphics[width=\patchwidth]{figures/defocus_map/#1_fusion_with_defocus_map_crop1.jpg}  \\
        &
        \includegraphics[width=\patchwidth]{figures/defocus_map/#1_source_crop2.jpg} &
        \includegraphics[width=\patchwidth]{figures/defocus_map/#1_warped_reference_crop2.jpg} &
        \includegraphics[width=\patchwidth]{figures/defocus_map/#1_fusion_no_defocus_map_crop2.jpg} &
        \includegraphics[width=\patchwidth]{figures/defocus_map/#1_fusion_with_defocus_map_crop2.jpg}  \\
        Full \wide and $\defocusmask$ & \wide & warped \tele & Ours w/o &  Ours  \\
        & & & $\defocusmask$ & \\
	}
    \begin{tabular}{ccccc}
        \addimage{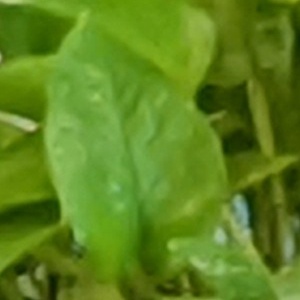}
    \end{tabular}
    \caption{\textbf{Contributions of the defocus map.} 
    We reject pixels in the white area of the defocus map from fusion.
    Using our defocus map, we preserve the details at fusion output on both defocused (top) and focused regions (bottom).
    }
    \label{fig:ablation_defocus_map}
\end{figure}

\begin{figure}
    \centering
    \footnotesize
    \renewcommand{\tabcolsep}{1pt} 
	\renewcommand{\arraystretch}{0.9} 
	\renewcommand{\imagewidth}{0.30\columnwidth} 
	\newcommand{\patchwidth}{0.18\columnwidth} 
	\newcommand{\verticaloffset}{1.38cm}
	\newcommand{\addimage}[1]{
        \multirow{3}{*}[\verticaloffset]{\includegraphics[width=\imagewidth]{figures/occlusion_map/#1_3_source_box.jpg}} &
        \multirow{3}{*}[\verticaloffset]{\includegraphics[width=\imagewidth]{figures/occlusion_map/#1_6_occlusion_mask_box.jpg}} &
        \includegraphics[width=\patchwidth]{figures/occlusion_map/#1_3_source_crop.jpg} &
        \includegraphics[width=\patchwidth]{figures/occlusion_map/#1_warped_reference_crop.jpg} \\
        & & \wide & Warped \tele \\
        & &
        \begin{overpic}[width=\patchwidth]{figures/occlusion_map/#1_7_fusion_no_occlusion_crop.jpg}
            \linethickness{0.7pt}
            \put(15,20){\color{green}\vector(1,0){30}}
            \put(30,80){\color{green}\vector(1,0){30}}
        \end{overpic} &
        \includegraphics[width=\patchwidth]{figures/occlusion_map/#1_7_fusion_crop.jpg} \\
        Full \wide & Full $\occlusionmask$ & Ours w/o $\occlusionmask$ &  Ours  \\
	}
    \begin{tabular}{cccc}
        \addimage{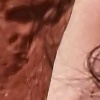}
    \end{tabular}
    \caption{\textbf{Contributions of the occlusion map.} We can reduce warping artifacts near occlusion boundaries with the help of occlusion map in fusion and blending.
    }
    \label{fig:ablation_occlusion_map}
\end{figure}

\subsection{Ablation study}

\begin{figure}
    \centering
    \footnotesize
    \renewcommand{\tabcolsep}{1pt} 
	\renewcommand{\arraystretch}{0.9} 
	\renewcommand{\imagewidth}{0.30\columnwidth} 
	\newcommand{\patchwidth}{0.18\columnwidth} 
	\newcommand{\verticaloffset}{1.38cm}
	\newcommand{\addimage}[1]{
        \multirow{3}{*}[\verticaloffset]{\includegraphics[width=\imagewidth]{figures/confidence_map/#1_source_box0.jpg}} &
        \multirow{3}{*}[\verticaloffset]{\includegraphics[width=\imagewidth]{figures/confidence_map/#1_confidence_map_box0.jpg}} &
        \includegraphics[width=\patchwidth]{figures/confidence_map/#1_source_crop0.jpg} &
        \includegraphics[width=\patchwidth]{figures/confidence_map/#1_warped_reference_crop0.jpg} \\
        & & \wide & Warped \tele \\
        & &
        \includegraphics[width=\patchwidth]{figures/confidence_map/#1_fusion_no_confidence_map_crop0.jpg} &
        \includegraphics[width=\patchwidth]{figures/confidence_map/#1_fusion_crop0.jpg} \\
        Full \wide & Full $\flowconfmask$ & Ours w/o $\flowconfmask$ &  Ours  \\
	}
    \begin{tabular}{cccc}
        \addimage{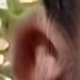}
    \end{tabular}
    \caption{\textbf{Contributions of the flow uncertainty map.} Optical flows are typically less robust on object boundaries, resulting in distortion and ghosting after fusion.
    }
    \label{fig:ablation_flow_uncertainty_map}
\end{figure}

\begin{figure}
    \centering
    \footnotesize
    \renewcommand{\tabcolsep}{1pt} 
	\renewcommand{\arraystretch}{0.9} 
	\renewcommand{\imagewidth}{0.246\columnwidth} 
	\newcommand{\patchwidth}{0.146\columnwidth} 
	\newcommand{\verticaloffset}{1.05cm}
	\newcommand{\addimage}[1]{
        \multirow{3}{*}[\verticaloffset]{\includegraphics[width=\imagewidth]{figures/rejection_map/#1_3_source_box.jpg}} &
        \multirow{3}{*}[\verticaloffset]{\includegraphics[width=\imagewidth]{figures/rejection_map/#1_15_rejection_map_box.jpg}} &
        \includegraphics[width=\patchwidth]{figures/rejection_map/#1_3_source_crop.jpg} &
        \includegraphics[width=\patchwidth]{figures/rejection_map/#1_4_reference_matched_color_crop.jpg} &
        \includegraphics[width=\patchwidth]{figures/rejection_map/#1_5_warped_reference_crop.jpg} \\
        & & \wide & \tele & Warped \tele \\
        & &
        \includegraphics[width=\patchwidth]{figures/rejection_map/#1_15_rejection_map_crop.jpg} &
        \begin{overpic}[width=\patchwidth]{figures/rejection_map/#1_7_fusion_no_rejection_crop.jpg}
            \linethickness{0.7pt}
            \put(95,95){\color{red}\vector(-1,-0.5){30}}
        \end{overpic} &
        \includegraphics[width=\patchwidth]{figures/rejection_map/#1_7_fusion_crop.jpg} \\
        Full \wide & Full $\rejectionmask$ & $\rejectionmask$ & Ours w/o &  Ours  \\
         &  &  & $\rejectionmask$ &   \\
	}
    \begin{tabular}{ccccc}
        \addimage{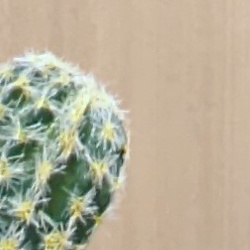}
    \end{tabular}
    \caption{\textbf{Contributions of the alignment rejection map.} Our alignment rejection map is able to identify mis-aligned pixels and remove ghosting artifacts from the fusion output.
    }
    \label{fig:ablation_rejection_map}
\end{figure}

\paragraph{Adaptive blending mask.}
We show the contributions of the defocus map, occlusion map, flow uncertainty map, and alignment rejection map in~\figref{ablation_defocus_map},~\ref{fig:ablation_occlusion_map},~\ref{fig:ablation_flow_uncertainty_map}, and~\ref{fig:ablation_rejection_map}.
Without the defocus map $\defocusmask$ in~\figref{ablation_defocus_map}, the background wall becomes blurrier than in the input \wide, as the blurry pixels in the defocused regions from \tele are fused.
Our defocus map excludes background pixels from fusion and preserves sharpness.
Without the occlusion mask in~\figref{ablation_occlusion_map}, misalignment on subject boundaries leads to visible artifacts on the fusion results.
As shown in~\figref{ablation_flow_uncertainty_map}, the flow uncertainty map identifies the regions with incorrect alignment and eliminates warping artifacts in the final output.
In~\figref{ablation_rejection_map}, the rejection map identifies misaligned pixels and avoids ghosting artifacts.

\begin{figure}
    \centering
    \footnotesize
    \renewcommand{\tabcolsep}{1pt} 
	\renewcommand{\arraystretch}{0.9} 
	\renewcommand{\imagewidth}{0.29\columnwidth}
	\newcommand{\patchwidth}{0.22\columnwidth}
	\newcommand{\addimage}[1]{
        \includegraphics[width=\imagewidth]{figures/vgg_loss/#1_source_box.jpg} &
        \includegraphics[width=\patchwidth]{figures/vgg_loss/#1_source_crop.jpg} &
        \begin{overpic}[width=\patchwidth]{figures/vgg_loss/#1_fusion_no_vgg_crop.jpg}
            \linethickness{0.7pt}
            \put(80,50){\color{red}\vector(0,-1){30}}
        \end{overpic} &
        \includegraphics[width=\patchwidth]{figures/vgg_loss/#1_fusion_with_vgg_crop.jpg} \\
        Full \wide & \wide & Ours w/o $\vggloss$ & Ours \\
	}
    \begin{tabular}{cccc}
        \addimage{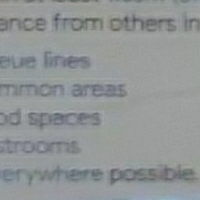}
    \end{tabular}
    \caption{\textbf{Effectiveness of the VGG loss.} VGG perceptual loss improves the sharpness and legibility of fusion results.
    }
    \label{fig:ablation_vgg_loss}
\end{figure}

\begin{figure}
    \centering
    \footnotesize
    \renewcommand{\tabcolsep}{1pt} 
	\renewcommand{\arraystretch}{0.9} 
	\renewcommand{\imagewidth}{0.19\columnwidth}
	\newcommand{\patchwidth}{0.25\columnwidth}
	\newcommand{\addimage}[1]{
        \includegraphics[width=\imagewidth]{figures/contextual_loss/#1_3_source_box.jpg} &
        \includegraphics[width=\patchwidth]{figures/contextual_loss/#1_3_source_crop.jpg} &
        \includegraphics[width=\patchwidth]{figures/contextual_loss/#1_7_fusion_no_contextual_crop.jpg} &
        \includegraphics[width=\patchwidth]{figures/contextual_loss/#1_7_fusion_crop.jpg} \\
        Full \wide & \wide & Ours w/o $\contextualloss$ & Ours \\
	}
    \begin{tabular}{cccc}
        \addimage{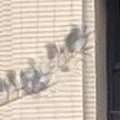}
    \end{tabular}
    \caption{\textbf{Effectiveness of the contextual loss.} Without contextual loss, results are blurry due to mis-aligned training data.
    }
    \label{fig:ablation_contextual_loss}
\end{figure}

\begin{figure}
    \centering
    \footnotesize
    \renewcommand{\tabcolsep}{1pt} 
	\renewcommand{\arraystretch}{0.9} 
	\renewcommand{\imagewidth}{0.30\columnwidth} 
	\newcommand{\patchwidth}{0.18\columnwidth} 
	\newcommand{\verticaloffset}{1.35cm}
	\newcommand{\addimage}[1]{
        \multirow{3}{*}[\verticaloffset]{\includegraphics[width=\imagewidth]{figures/gaussian_loss/#1_source_box.jpg}} &
        \multirow{3}{*}[\verticaloffset]{\includegraphics[width=\imagewidth]{figures/gaussian_loss/#1_warped_reference_box.jpg}} &
        \includegraphics[width=\patchwidth]{figures/gaussian_loss/#1_source_crop.jpg} &
        \includegraphics[width=\patchwidth]{figures/gaussian_loss/#1_warped_reference_crop.jpg} \\
        & & \wide & Warped \tele \\
        & &
        \begin{overpic}[width=\patchwidth]{figures/gaussian_loss/#1_fusion_no_gaussian_crop.jpg}
            \linethickness{0.7pt}
            \put(45,20){\color{red}\vector(1,0){30}}
            \put(45,80){\color{red}\vector(1,0){30}}
            \put(10,15){\color{red}\vector(1,1){20}}
        \end{overpic} &
        \includegraphics[width=\patchwidth]{figures/gaussian_loss/#1_fusion_with_gaussian_crop.jpg} \\
        Full \wide & Full warped \tele & Ours w/o &  Ours  \\
         &  & $\colorloss$ &   \\
	}
    \begin{tabular}{cccc}
        \addimage{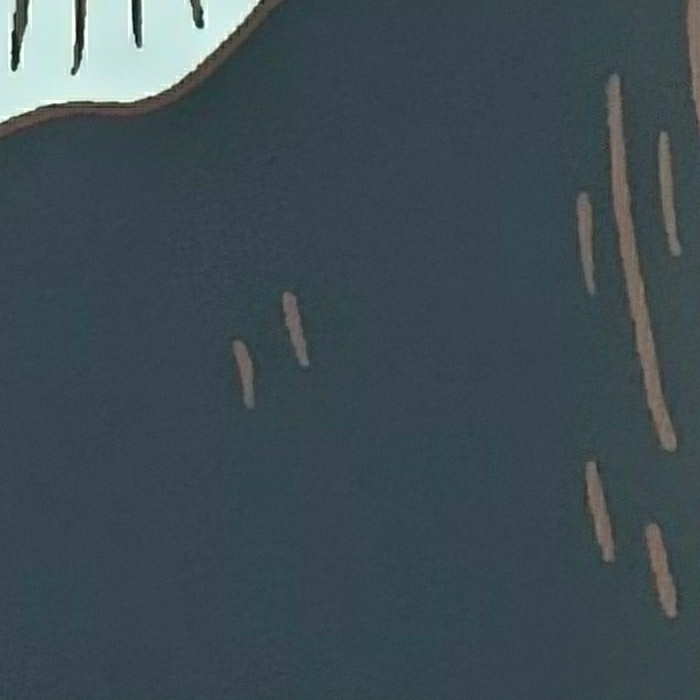}
    \end{tabular}
    \caption{\textbf{Contributions of the brightness consistency loss.} Without the brightness consistency loss, our fusion result shows inconsistent colors on the bear's body (please zoom in to see the details)
    }
    \label{fig:ablation_gaussian_loss}
\end{figure}

\paragraph{Training losses.}
We evaluate the contribution of perceptual~(\eqnref{vgg_loss}), contextual loss~(\eqnref{contextual_loss}), and brightness consistency loss~(\eqnref{color_loss}).
In~\figref{ablation_vgg_loss}, the VGG loss helps to recover sharper details and more visually pleasing results.
The contextual loss~\cite{mechrez2018contextual} in~(\eqnref{contextual_loss}) minimizes the differences in the semantic feature space and relaxes the pixel alignment constraints, which helps to train our Fusion UNet on the dual-rig dataset.
In~\figref{ablation_contextual_loss}, without the contextual loss, Fusion UNet generates blurry pixels when the \wide and \tele are not well aligned.
As shown in~\figref{ablation_gaussian_loss}, with the color-consistency loss, our model can preserve the original color of \wide on the fusion result and be robust to the auto-exposure metering mismatch between \wide and \tele cameras.

\begin{figure}
    \centering
    \footnotesize
    \renewcommand{\tabcolsep}{1pt} 
	\renewcommand{\arraystretch}{1} 
	\renewcommand{\imagewidth}{0.47\columnwidth} 
	\newcommand{\patchwidth}{0.165\columnwidth} 
	\newcommand{\verticaloffset}{1.23cm}
    \begin{tabular}{cccc}
        \multirow{2}{*}[\verticaloffset]{\includegraphics[width=\imagewidth]{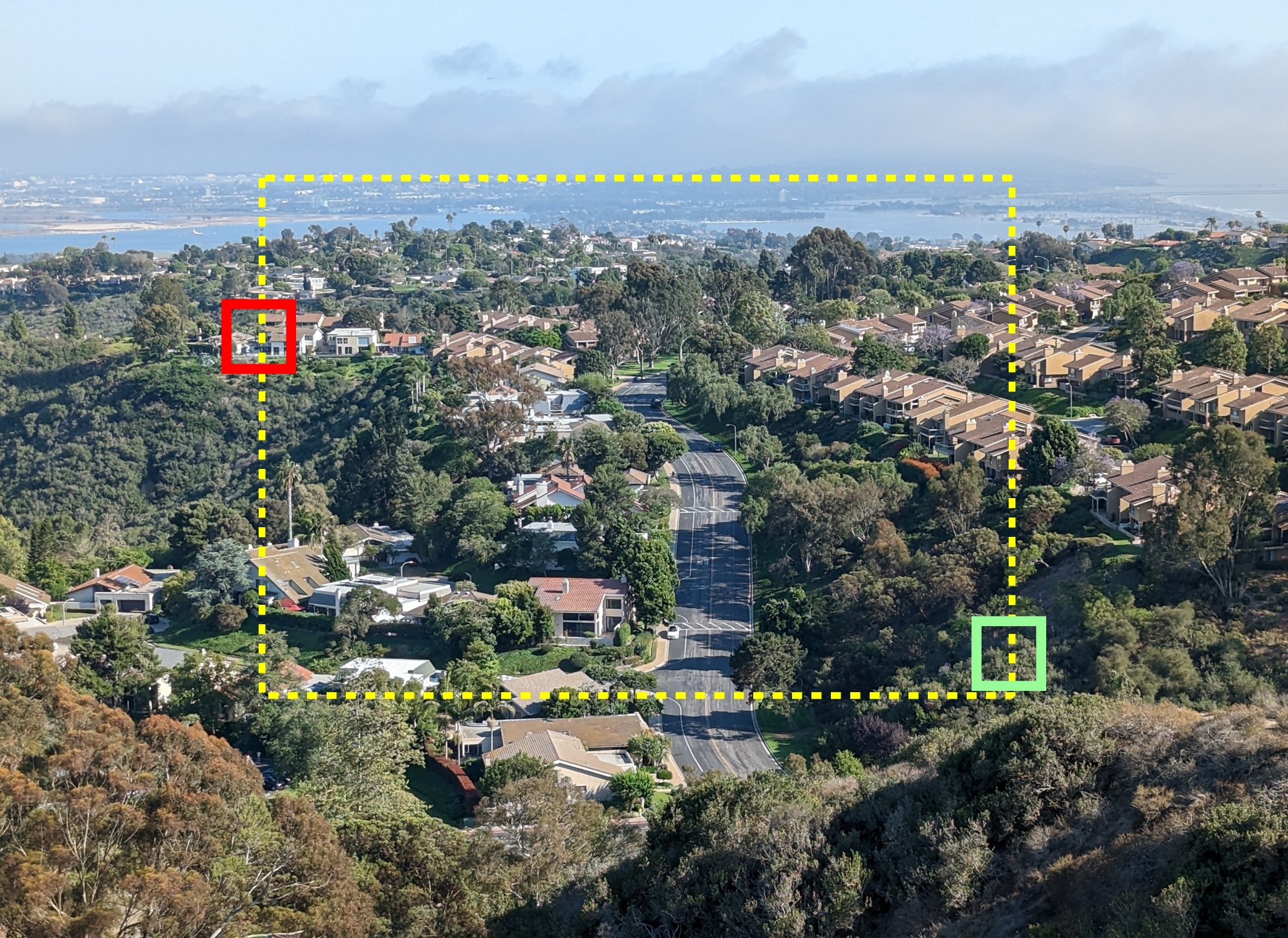}} &
        \includegraphics[width=\patchwidth]{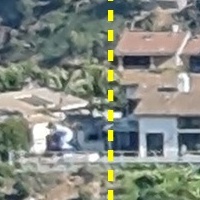} &
        \includegraphics[width=\patchwidth]{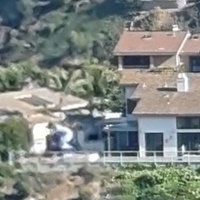} &
        \includegraphics[width=\patchwidth]{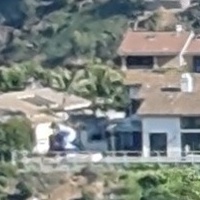} \\
        & 
        \includegraphics[width=\patchwidth]{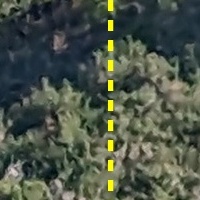} &
        \includegraphics[width=\patchwidth]{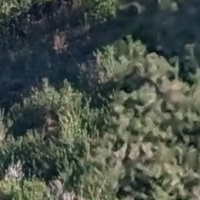} &
        \includegraphics[width=\patchwidth]{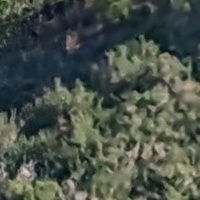} \\
        Full \wide & \wide & Ours w/o  &  Ours  \\
        &  & smoothing & 
    \end{tabular}
    \caption{\textbf{Effectiveness of boundary smoothing.}  The yellow dotted lines in (a) show the fusion ROI (i.e., FOV of \tele).
    With boundary smoothing, the fusion boundary looks more natural and smooth.
    }
    \label{fig:fusion_transition}
\end{figure}

\paragraph{Fusion boundary.}
When blending the fusion output back to the full \wide image, we apply Gaussian smoothing to smooth the blending boundary to avoid abrupt transitions.
Without boundary smoothing, we can see invariant details on the building and trees in~\figref{fusion_transition}(b).
While boundary smoothing sacrifices some detail improvements around the transition boundary, our results in~\figref{fusion_transition}(c) look more natural after blending.

\subsection{Limitations}
\begin{figure}
    \centering
    \footnotesize
    \renewcommand{\tabcolsep}{1pt} 
	\renewcommand{\arraystretch}{0.9} 
	\renewcommand{\imagewidth}{0.19\columnwidth}
	\newcommand{\patchwidth}{0.25\columnwidth}
	\newcommand{\addimage}[1]{
        \includegraphics[width=\imagewidth]{figures/low_light_scene/#1_LR_box.jpg} &
        \includegraphics[width=\patchwidth]{figures/low_light_scene/#1_LR_crop.jpg} &
        \begin{overpic}[width=\patchwidth]{figures/low_light_scene/#1_Ref_crop.jpg}
        \end{overpic} &
        \includegraphics[width=\patchwidth]{figures/low_light_scene/#1_crop} \\
        Full \wide & \wide & \tele & Ours (force fusion)
	}
    \begin{tabular}{cccc}
        \addimage{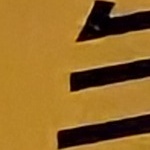}
    \end{tabular}
    \caption{\textbf{Limitation on low-light.} Under extremely low-light condition, \tele becomes too noisy. Our fusion will transfer noise to the output image in such a case. Therefore, we design our system to skip fusion based on \tele SNR.
    }
    \label{fig:low_light_limitation}
\end{figure}

\begin{figure}
    \centering
    \footnotesize
	\renewcommand{\tabcolsep}{1pt} 
	\renewcommand{\arraystretch}{1} 
	\renewcommand{\imagewidth}{0.265\columnwidth} 
	\newcommand{\patchwidth}{0.139\columnwidth} 
	\newcommand{\verticaloffset}{1.0cm}
	\newcommand{\addimage}[1]{
        \multirow{3}{*}[\verticaloffset]{\includegraphics[width=\imagewidth]{figures/non_overlap_fov/#1_LR_box.jpg}} &
        \multirow{3}{*}[\verticaloffset]{\includegraphics[width=\imagewidth]{figures/non_overlap_fov/#1_Ref_box.jpg}} &
        \includegraphics[width=\patchwidth]{figures/non_overlap_fov/#1_LR_crop1.jpg} &
        \includegraphics[width=\patchwidth]{figures/non_overlap_fov/#1_FusionZoom_crop1.jpg} & 
        \includegraphics[width=\patchwidth]{figures/non_overlap_fov/#1_DCSR_crop1.jpg} \\
        & & \wide & Ours & DCSR \\
        & & \multicolumn{3}{c}{(Within \tele FOV)} \\
        & &
        \includegraphics[width=\patchwidth]{figures/non_overlap_fov/#1_LR_crop0.jpg} &
        \includegraphics[width=\patchwidth]{figures/non_overlap_fov/#1_FusionZoom_crop0.jpg} &
        \includegraphics[width=\patchwidth]{figures/non_overlap_fov/#1_DCSR_crop0.jpg}\\
        Full \wide & Full \tele & \wide & Ours & DCSR \\
        & & \multicolumn{3}{c}{(Outside \tele FOV)}
	}
    \begin{tabular}{ccccc}
        \addimage{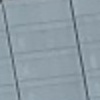}
    \end{tabular}
    \caption{\textbf{Limitation on non-overlapping FOV.} For the pixels outside \tele FOV, our method maintains the same values as \wide, while DCSR is able to enhance some details.}
    \label{fig:non_matching_fov_scene}
\end{figure}

Our system has the following limitations: 
1) Under extremely low-light conditions (less than $5$ lux), the \tele image becomes too noisy due to sensor limitations, as shown in~\figref{low_light_limitation}.
2) If the synchronization between \tele and \wide exceeds a limit (e.g., 128ms in our system), the alignment will be very challenging, and our system will skip fusion to prevent alignment artifacts.
3) Our system does not enhance the details outside the FOV of \tele, while existing methods (e.g., DCSR~\cite{wang2021dual}) can improve details on the whole image via learning SISR or finding long-range correspondences, as shown in~\figref{non_matching_fov_scene}.

\section{Conclusions}

In this work, we present a robust system for hybrid zoom super-resolution on mobile devices. 
We develop efficient ML models for alignment and fusion, propose an adaptive blending algorithm to account for imperfections in real-world images, and design a training strategy using an auxiliary camera to minimize domain gaps.
Our system achieves an interactive speed (500 ms to process a 12MP image) on mobile devices and is competitive against state-of-the-art methods on public benchmarks and our \dataset{Hzsr} dataset.

\begin{acks}
This work would be impossible without close collaboration between many teams within Google.
We are particularly grateful to Li-Chuan Yang, Sung-Fang Tsai, Gabriel Nava, Ying Chen Lou, and Lida Wang for their work on integrating the dual camera system.
We also thank Junhwa Hur, Sam Hasinoff, Mauricio Delbracio, and Peyman Milanfar for their advice on algorithm development.
We are grateful to Henrique Maia and Brandon Fung for their helpful discussions on the paper. 
Finally, we thank Hsin-Fu Wang, Lala Hsieh, Yun-Wen Wang, Daniel Tat, Alexander Tat and all the photography models for their significant contributions to data collection and image quality reviewing. 
\end{acks}


\clearpage
\bibliographystyle{ACM-Reference-Format}
\bibliography{references}

\end{document}